%% file: main.tex
\documentclass[10pt,twocolumn,letterpaper]{article}

\usepackage[pagenumbers]{cvpr} 
\usepackage{multirow}
\input{preamble}
\definecolor{cvprblue}{rgb}{0.21,0.49,0.74}
\usepackage[pagebackref,breaklinks,colorlinks,allcolors=cvprblue]{hyperref}

\def\paperID{22027} 
\def\confName{CVPR}
\def\confYear{2026}

\title{NanoSD: Edge Efficient Foundation Model for Real Time Image Restoration}

\author{
Subhajit Sanyal$^{1,*}$ \and
Srinivas Soumitri Miriyala$^{1,*}$ \and
Akshay Janardan Bankar$^{1,*}$ \and
Manjunath Arveti$^{1}$ \and
Sowmya Vajrala$^{1}$ \and
Shreyas Pandith$^{1}$ \and
Sravanth Kodavanti$^{1}$ \and
Abhishek Ameta$^{1}$ \and
Harshit$^{1}$ \and
Amit Satish Unde$^{1}$ \and
\\
$^{1}$Samsung Research India, Bangalore \\
{\tt\small
\{subhajit.sn, srinivas.m1, a.bankar, at.manjunath, v.lahari, s.pandith,}
\\
{\tt\small k.sravanth, abhishek.amt, harshit.5, amit.unde\}@samsung.com}
}

\begin{document}

\twocolumn[{%
\renewcommand\twocolumn[1][]{#1}%
\maketitle
\begin{center}
    \centering
    \captionsetup{type=figure}
    \includegraphics[width=1\linewidth]{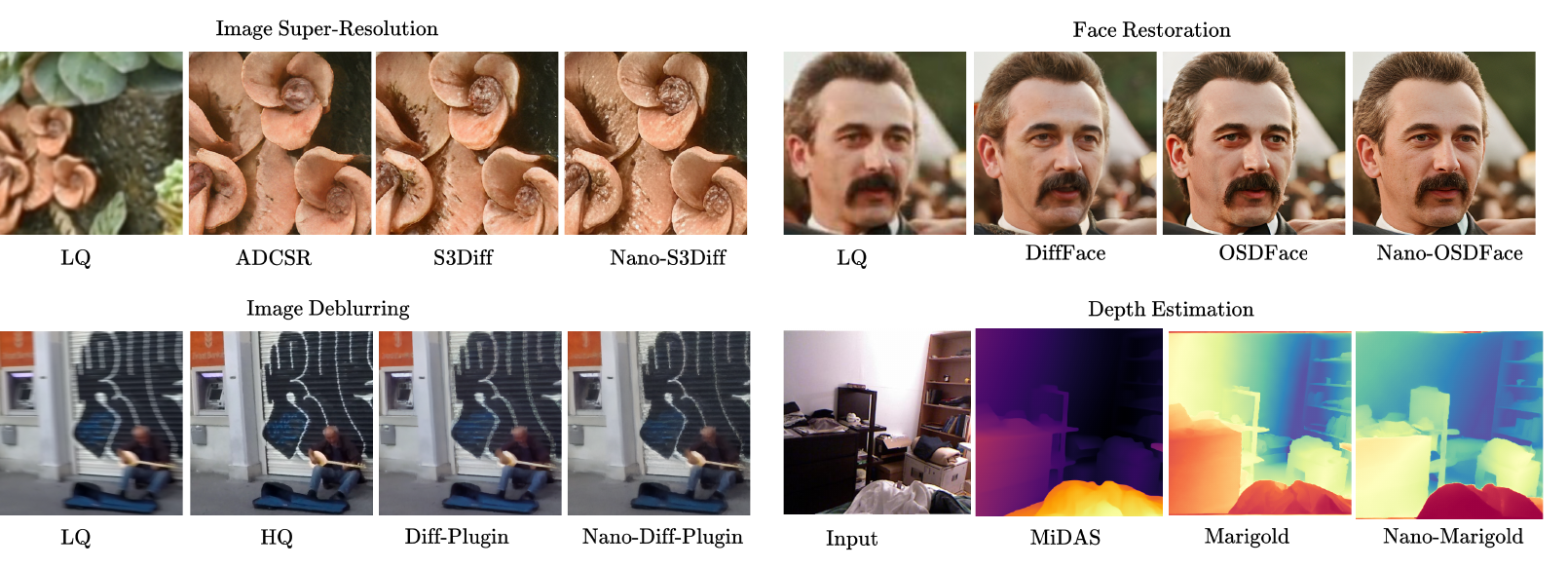}
    \captionof{figure}{\textbf{We present NanoSD, an edge-efficient diffusion model designed for image restoration applications.} Its core principle is to harness the extensive visual priors embedded in contemporary generative image models while maintaining computational feasibility for edge deployment. NanoSD demonstrates practical applicability across diverse low-level vision tasks in real-world scenarios.}
\end{center}%
}]

\maketitle
\renewcommand{\thefootnote}{\fnsymbol{footnote}}
\footnotetext[1]{Equal contribution}

\input{sec/0_abstract}    
\input{sec/introduction1}

\input{sec/RelatedWork}

\input{sec/method}

\input{sec/2_formatting_2}
\input{sec/conclusion}
{
    \small
    \bibliographystyle{ieeenat_fullname}
    \bibliography{main}
}
\input{sec/X_suppl}


\end{document}

%% file: sec/0_abstract.tex
\begin{abstract}
Latent diffusion models such as Stable Diffusion 1.5 offer strong generative priors that are highly valuable for image restoration, yet their full pipelines remain too computationally heavy for deployment on edge devices. Existing lightweight variants predominantly compress the denoising U-Net or reduce the diffusion trajectory, which disrupts the underlying latent manifold and limits generalization beyond a single task. We introduce NanoSD, a family of Pareto-optimal diffusion foundation models distilled from Stable Diffusion 1.5 through network surgery, feature-wise generative distillation, and structured architectural scaling jointly applied to the U-Net and the VAE encoder–decoder. This full-pipeline co-design preserves the generative prior while producing models that occupy distinct operating points along the accuracy–latency–size frontier (e.g., 130M–315M parameters, achieving real-time inference down to 20ms on mobile-class NPUs). We show that parameter reduction alone does not correlate with hardware efficiency, and we provide an analysis revealing how architectural balance, feature routing, and latent-space preservation jointly shape true on-device latency. When used as a drop-in backbone, NanoSD enables state-of-the-art performance across image super-resolution, image deblurring, face restoration, and monocular depth estimation, outperforming prior lightweight diffusion models in both perceptual quality and practical deployability. NanoSD establishes a general-purpose diffusion foundation model family suitable for real-time visual generation and restoration on edge devices.
\end{abstract}

%% file: sec/introduction1.tex
\section{Introduction}
\label{sec:introduction}

Image restoration (IR) focuses on reconstructing high-quality (HQ) images from their degraded low-quality (LQ) counterparts \cite{sd, diffplugin, codeformer}. Recent advancements in diffusion model (DM) acceleration have garnered significant interest, particularly in the context of image super-resolution (SR) \cite{sinsr, adcsr, edgesdsr}. While several one-step diffusion-based approaches have demonstrated promising results, their computational complexity remains prohibitive for real-world deployment \cite{sinsr, osediff, s3diff}. To address these challenges, recent research has prioritized architectural optimizations to reduce redundancy in large-scale models \cite{snapfusion, snapgen}. Although these methods maintain strong generation capabilities for text-to-image (T2I) tasks, their practical deployment on computationally limited platforms faces substantial hurdles due to excessive model sizes and latency \cite{snapfusion, mobilediffusion}. These limitations are further compounded in IR, which inherently operates on high-resolution inputs and often requires multiple model executions \cite{edgesdsr}.

To enable the practical deployment of diffusion-based restoration on edge devices, several distillation techniques have been explored, particularly for SR tasks \cite{tinysr, pocketsr, edgesdsr}. While these methods demonstrate promising efficiency gains, they are primarily optimized for SR tasks using limited datasets, preventing them from fully leveraging the rich prior knowledge encoded in large pre-trained T2I models. Consequently, such distilled models often produce structurally implausible results with perceptually suboptimal details. Additionally, existing methods lack the flexibility to accommodate diverse conditioning strategies required for broader IR tasks \cite{adcsr, edgesdsr, pocketsr, tinysr}. For example, S3Diff \cite{s3diff} incorporates a degradation-aware Low-Rank Adaptation (LoRA) module, while OSDFace \cite{osdface} utilizes visual prompts as supplementary conditioning. Similarly, Diff-Plugin \cite{diffplugin} introduces a lightweight task-specific module via cross-attention mechanisms, and DiffBIR \cite{diffbir} employs a ControlNet-based \cite{controlnet} architecture for image conditioning. These specialized approaches highlight the need for a versatile framework for edge-efficient diffusion methods to support a wider range of restoration scenarios.

 

Designing edge-efficient diffusion models for diverse restoration tasks hinges on effectively integrating conditional information to guide accurate and natural outputs—an aspect often overlooked in prior work. Also, since modern diffusion models are pretrained on large-scale datasets, preserving their latent space is crucial. This paper introduces NanoSD, a latent diffusion model based on the Stable Diffusion 1.5 (SD~1.5) \cite{sd}, optimized for edge efficiency while ensuring compatibility with various control plugins. Our contributions in this paper are as follows:

\begin{itemize}
    \item We introduce a hardware-aware reformulation of the SD~1.5 U-Net by decomposing it into stage-wise dimensions and constructing a compact set of shape-preserving block variants tailored to edge accelerators.
    
    \item We propose a block-level generative distillation strategy that aligns each candidate block variant with its SD~1.5 teacher counterpart, enabling large-scale architectural exploration without full-model retraining.
    
    \item We cast the selection of an efficient diffusion backbone as a multi-objective optimization problem over teacher-aligned Fr\'echet Inception Distance (FID), on-device latency, and parameter count, and employ Bayesian optimization to obtain a Pareto set of U-Nets capturing diverse accuracy--efficiency trade-offs.
    
    \item From this Pareto frontier, we identify a balanced architecture---\emph{NanoSD}---and distill the corresponding VAE encoder and decoder to construct a fully lightweight latent diffusion pipeline.
    
    \item Through extensive experiments across super-resolution, face restoration, deblurring, dehazing, deraining, desnowing, and monocular depth estimation, we demonstrate that NanoSD maintains the generative behavior of SD~1.5 while substantially reducing computational cost, enabling practical deployment on mobile-class NPUs.
\end{itemize}


%% file: sec/RelatedWork.tex
\section{Related Work}
\label{sec:literature}
Significant progress in diffusion models, particularly for text-to-image (T2I) synthesis (e.g., SD \cite{sd}, SD3 \cite{sd3}), has spurred the development of pre-trained diffusion-based image restoration techniques such as StableSR \cite{stablesr}, Diff-Plugin \cite{diffplugin}, DiffPIR \cite{diffpir}, DiffBIR \cite{diffbir}, PASD \cite{pasd}, and SeeSR \cite{seesr}. These methods leverage the robust priors of T2I models, but suffers from substantial latency and computational overhead. To mitigate this, recent efforts focus on single-step frameworks, primarily targeting SR. For instance, SinSR \cite{sinsr} accelerates ResShift \cite{resshift} via bidirectional distillation, while OSEDiff \cite{osediff} employs variational score distillation approach to enhance the realism of super-resolved outputs. Building on distilled SD-Turbo architectures, S3Diff \cite{s3diff} integrates a degradation-guided LoRA module to improve SR quality. Albeit promising, high computational cost of these networks hinder their practical deployment on resource-constrained devices.

Recent research in efficient diffusion models has prioritized architectural optimizations. SnapFusion \cite{snapfusion} analyzes individual module contributions to balance efficiency and accuracy, while MobileDiff \cite{mobilediffusion} relocates Transformer blocks to lower-resolution stages for computational savings. SnapGen \cite{snapgen} further reduces model size and computation by eliminating high-resolution attention and replacing standard convolutions with depthwise separable variants. Deployment of these methods for high-resolution image restoration remains challenging due to latency constraints. To enhance edge efficiency for SR, AdcSR \cite{adcsr} leverages adversarial diffusion compression to distill OSEDiff \cite{osediff}, significantly reducing complexity by pruning the VAE encoder and denoising UNet. Edge-SD-SR \cite{edgesdsr} introduces a novel conditioning mechanism for low-resolution inputs, while TinySR \cite{tinysr} employs depthwise UNet pruning and conditional information removal to improve efficiency. PocketSR \cite{pocketsr} adopts multi-layer feature distillation for enhanced knowledge transfer. However, current edge-efficient models underutilize the generative capabilities of T2I frameworks and lack modular support for advanced conditioning mechanisms (e.g., LoRA \cite{osediff, s3diff}, ControlNet \cite{diffbir}), limiting their applicability to diverse restoration tasks \cite{diffplugin, osdface}.

%% file: sec/method.tex
\section{Method}
\label{sec:method}

\begin{figure*}
    \centering
    \includegraphics[width=1\linewidth]{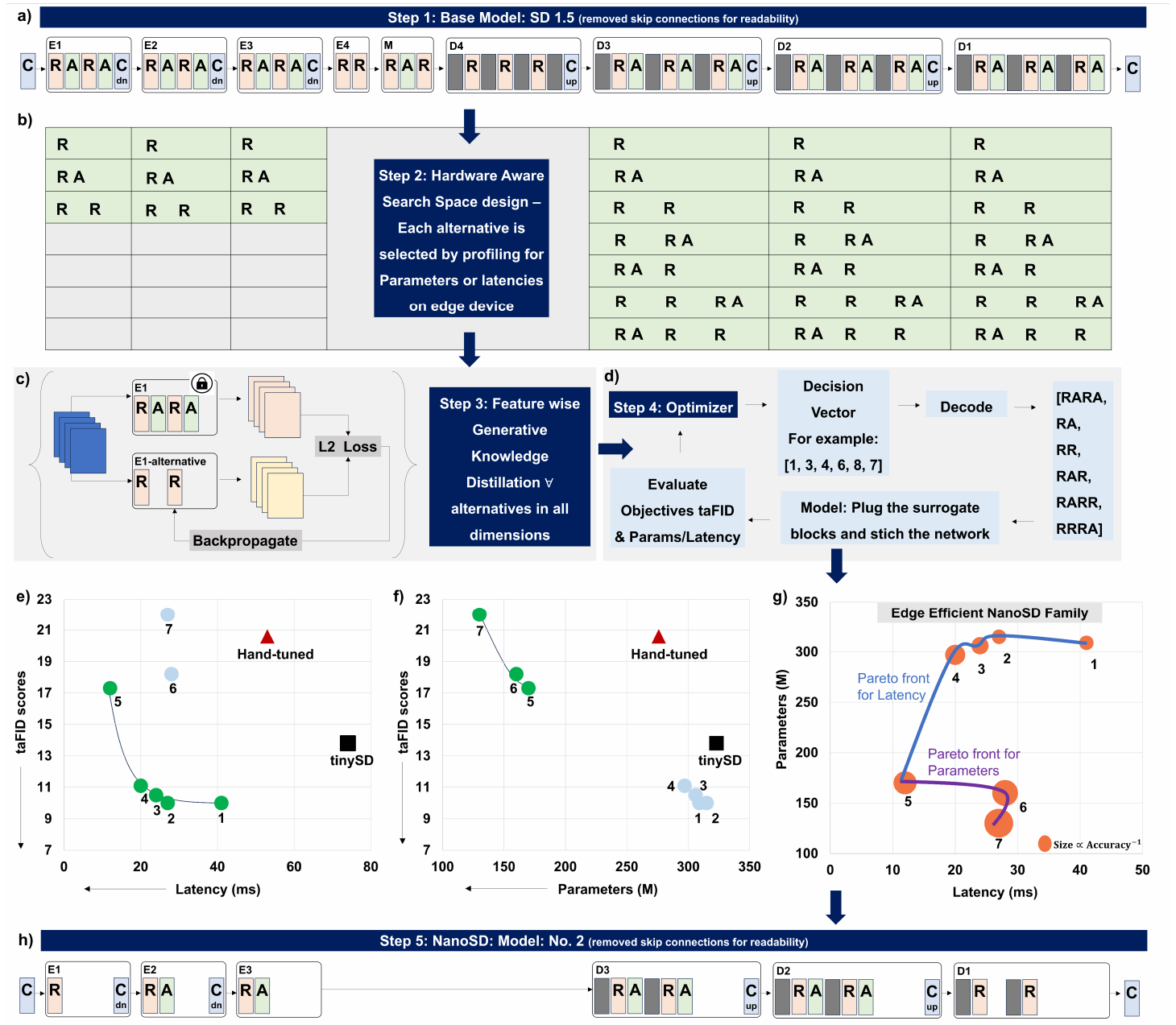}
    \caption{
    \textbf{Overview of the NanoSD framework.}  
    (a) Baseline SD~1.5 U\textendash Net architecture, shown with skip connections removed for readability.  
    (b) Hardware-aware search space construction: for each of the six retained stages (three encoders and three decoders), we derive shape-preserving residual/attention variants that are profiled on the target edge device for latency and parameter cost.  
    (c) Feature-wise generative distillation: each candidate block is distilled independently from its corresponding SD~1.5 teacher block using an $\ell_2$ feature-matching loss.  
    (d) Combinatorial assembly and evaluation: a decision vector specifies one distilled block per stage, producing a structurally valid U\textendash Net; each assembled model is evaluated on taFID and either on-device latency or parameter count via Bayesian optimization.  
    (e) Latency--taFID Pareto frontier obtained from the search; green points denote Pareto-optimal models, and non-Pareto points are shown for comparison.  
    (f) Parameter--taFID Pareto frontier.  
    Both plots also include a hand-tuned baseline and the Segmind TinySD model, which fall far from the frontier.  
    (g) The resulting \emph{NanoSD family}: seven Pareto-optimal architectures spanning different accuracy--efficiency trade-offs (bubble size inversely proportional to accuracy).  
    (h) Final selected architecture, \emph{NanoSD} (Model~2), used for all downstream experiments; skip connections are omitted for clarity.  
    }
    \label{fig:nas}
\end{figure*}

Our objective is to derive an edge-efficient diffusion foundation model from SD~1.5 baseline \cite{sd}. Although SD~1.5 provides strong generative priors, its U-Net and VAE components remain prohibitively heavy for embedded NPUs. Moreover, theoretical compute metrics (FLOPs/GMACs) do not reliably predict practical latency, since NPUs are optimized for specific operator patterns (e.g., GEMM), tensor layouts, and memory behavior. As a result, reducing arithmetic compute alone does not guarantee proportional latency improvements. To address this mismatch, we introduce a hardware-aware architectural redesign and distillation framework (Fig.~\ref{fig:nas}) that restructures SD~1.5 at the block level, constructs a tractable search space aligned with edge hardware (Fig.~\ref{fig:nas}b), trains locally distilled surrogate blocks that preserve SD~1.5’s generative behavior (Fig.~\ref{fig:nas}c), and selects full U-Net architectures using multi-objective Bayesian optimization (Fig.~\ref{fig:nas}d). This search over the distilled block variants produces a Pareto-optimal set of U-Net backbones, each representing a different accuracy–efficiency trade-off. From this set, we select a single U-Net that offers the most balanced operating point and fix it as our base architecture \emph{NanoSD}. Keeping this U-Net frozen, we then distill the corresponding VAE encoder and decoder from SD~1.5, yielding a fully lightweight latent diffusion pipeline tailored to the selected U-Net.

\subsection{Hardware-Aware Decomposition of U-Net}

The SD~1.5 U-Net consists of four encoders, a middle block, and four decoders. Prior analyses in lightweight diffusion literature---including SnapFusion \cite{snapfusion}, Segmind TinySD \cite{tinysd}, and Koala \cite{koala}---report that the deepest encoder (encoder-4), the adjacent middle block, and the corresponding decoder-4 contribute minimally to generative fidelity. Motivated by these findings, we remove these three stages entirely from our design space. The remaining backbone comprises three encoders and three decoders, each treated as an independent architectural dimension.

For each stage, we inspect the original SD~1.5 block structure and derive hardware-friendly variants that strictly preserve input/output tensor shapes. For example, the SD~1.5 block in the first decoder stage contains a sequence of residual and attention modules followed by a convolution, denoted R--A--R--A. We construct shape-preserving variants (e.g., residual-only or reduced-attention configurations) such that all alternatives remain strictly compatible with upstream and downstream tensors. Applying this procedure to all six stages yields the full combinatorial design space illustrated in Fig.~\ref{fig:nas}a and Fig.~\ref{fig:nas}b. The resulting space consists of 32,768 ($4 \times 4 \times 4 \times 8 \times 8 \times 8$) candidate U-Net architectures, each of which is dimensionally valid without requiring adapters or tensor reshaping.

\begin{table*}[!h]
\centering
\caption{
\textbf{Comparison of TinySD (Segmind), hand-tuned baseline, and NanoSD variants.}
Latency is measured on a Qualcomm NPU using 8-bit weights and 16-bit activations.  
taFID denotes the teacher-aligned FID metric used during the search.  
The seven NanoSD models correspond to all Pareto-optimal architectures obtained from the latency--taFID and parameter--taFID objectives.  
For each model, we list the selected block variants across the retained stages of the SD~1.5 U\textendash Net (E1--E3 for encoders and D3--D1 for decoders).  
Model~5 achieves the lowest latency, Model~7 achieves the fewest parameters, and Model~2 provides the best overall balance of accuracy and efficiency and is used as \emph{NanoSD} in all downstream experiments.
}
\label{table:nas_arch}
\begin{tabular}{|l|l|l|l|l|l|l|l|l|l|}
\hline
\textbf{Model}      & \textbf{Latency (ms)} & \textbf{Parameters (M)} & \textbf{taFID} & \textbf{E1} & \textbf{E2} & \textbf{E3} & \textbf{D3} & \textbf{D2} & \textbf{D1} \\
\hline
\textbf{TinySD}     & 74                    & 323                     & 13.8           & RA          & RA          & RA          & RA          & RA          & RA          \\
\hline
\textbf{Hand-tuned} & 53                    & 276                     & 20.6           & RA          & RA          & RA          & RAR         & RAR         & RAR         \\
\hline
\textbf{NanoSD 1}   & 41                    & 309                     & 10             & R           & RA          & RA          & RARA        & RRA         & RRA         \\
\hline
\textbf{NanoSD 2}   & 27                    & 315                     & 10             & R           & RA          & RA          & RARA        & RARA        & RR          \\
\hline
\textbf{NanoSD 3}   & 24                    & 306                     & 10.5           & R           & RA          & RA          & RARA        & RRA         & RR          \\
\hline
\textbf{NanoSD 4}   & 20                    & 297                     & 11.1           & R           & RA          & RA          & RARA        & RR          & RR          \\
\hline
\textbf{NanoSD 5}   & 12                    & 170                     & 17.3           & R           & R           & R           & RA          & RR          & RR          \\
\hline
\textbf{NanoSD 6}   & 28                    & 160                     & 18.2           & R           & R           & R           & RA          & RA          & RA          \\
\hline
\textbf{NanoSD 7}   & 27                    & 130                     & 22             & R           & R           & R           & R           & RA          & RRA         \\
\hline
\end{tabular}
\end{table*}

\subsection{Feature-Wise Generative Distillation}

Training all 32,768 architectures end-to-end is infeasible. We therefore adopt a divide-and-conquer distillation strategy at the block level. For a given stage $i$ and candidate variant $j$, let $\mathcal{B}_{i}$ denote the SD~1.5 block (teacher) and $\mathcal{B}_{i,j}$ the corresponding variant (student). For input features $F$, the teacher and student outputs are
\begin{equation}
O_T = \mathcal{B}_i(F), \qquad 
O_S = \mathcal{B}_{i,j}(F).
\label{eq:block_outputs}
\end{equation}
Each variant is trained with the feature-matching objective
\begin{equation}
\mathcal{L}^{(i,j)}_{\mathrm{distill}}
= \| O_S - O_T \|_2^2.
\label{eq:distill_loss}
\end{equation}
Since all blocks are distilled independently, this process is massively parallel and computationally lightweight. Across the six stages, we obtain 30 (3+3+3+7+7+7) distilled surrogate blocks (see Supplementary Material for details on training). This preserves SD~1.5’s local generative behavior, enabling full U-Nets assembled from these surrogates to approximate SD~1.5 without full-network training.

\subsection{Assembling Candidate Diffusion U-Nets}

A complete U-Net architecture is encoded by a discrete vector $\mathbf{z} = [z_1, z_2, z_3, z_4, z_5, z_6]$, where each element selects one surrogate block for the corresponding stage. Because all variants maintain spatial and channel compatibility, substituting blocks according to $\mathbf{z}$ yields a structurally valid U-Net. Combined with the original SD~1.5 VAE encoder and decoder, this produces a functional latent diffusion pipeline, evaluated using the procedure outlined in Fig.~\ref{fig:nas}d.

\subsection{Optimization in the H/W-Aligned Space}

To measure fidelity, we construct a test dataset of prompts and corresponding SD~1.5 outputs. For any architecture $\mathbf{z}$, we generate outputs $\hat{X}(\mathbf{z})$ using the same prompts and compute the FID:
\begin{equation}
f_{\mathrm{FID}}(\mathbf{z}) 
= \mathrm{FID}\!\left( \hat{X}(\mathbf{z}), X_{\mathrm{SD1.5}} \right),
\label{eq:fid}
\end{equation}
which quantifies deviation from SD~1.5’s generative distribution. We report this as a teacher-aligned Fréchet Inception Distance (taFID), as it measures the distributional divergence between the distilled model and the SD~1.5 teacher using identical prompt–seed pairs, serving as a relative fidelity metric rather than a canonical dataset FID.

To assess efficiency, we consider both parameter count and on-device latency as separate second objectives:
\begin{equation}
f_{\mathrm{param}}(\mathbf{z}), \qquad 
f_{\mathrm{latency}}(\mathbf{z}).
\label{eq:efficiency_objs}
\end{equation}
This yields two bi-objective problems:
\begin{equation}
\min_{\mathbf{z}} \; 
\left( f_{\mathrm{FID}}(\mathbf{z}), \; f_{\mathrm{param}}(\mathbf{z}) \right),
\label{eq:biobj_param}
\end{equation}
\begin{equation}
\min_{\mathbf{z}} \; 
\left( f_{\mathrm{FID}}(\mathbf{z}), \; f_{\mathrm{latency}}(\mathbf{z}) \right),
\label{eq:biobj_latency}
\end{equation}
with $\mathbf{z}$ constrained to the discrete space defined by Fig. \ref{fig:nas}b.

Because FID evaluation remains costly even without training, we adopt Bayesian Optimization (BO). We relax $\mathbf{z}$ to a continuous representation $\mathbf{x} \in [0,1]^6$ and project it to the nearest feasible architecture via a deterministic mapping. Gaussian Process models approximate both objectives, and candidate points are selected by maximizing Expected Hypervolume Improvement (EHVI). Full derivations and analyses are provided in the Supplementary Material. BO yields a sparse Pareto frontier of architectures balancing generative fidelity and hardware efficiency.

\subsection{VAE Distillation and End-to-End Training}

After selecting Pareto-optimal U-Nets, we freeze their weights and distill both the VAE encoder and decoder to reduce overall pipeline size. VAE distillation follows standard feature-matching losses between SD~1.5 teacher outputs and student predictions (details in Supplementary Material). Each resulting pair of distilled U-Net and distilled VAE forms a NanoSD candidate.

To correct accumulated discrepancies from block-level distillation, we fine-tune each NanoSD model using the standard diffusion denoising objective. The resulting NanoSD family is evaluated across super-resolution, face restoration, deblurring, dehazing, deraining, desnowing, and monocular depth estimation, demonstrating that the distilled architectures retain SD~1.5’s generative versatility while enabling practical deployment on edge devices.

%% file: sec/2_formatting_2.tex
\begin{table*}[]
\centering
\caption{Quantitative comparison of different methods on DIV-2K Val \cite{div2kval} dataset. The best, second-best and third-best results are highlighted in red, blue, and green colors, respectively. Refer to Supplemental for detailed analysis.}
\label{table:sr1}
\begin{tabular}{|l|l|l|l|l|l|l|l|l|l|}
\hline
Method                     & PSNR$\uparrow$  & SSIM$\uparrow$   & LPIPS$\downarrow$   & FID$\downarrow$  & NIQE$\downarrow$ & MUSIQ$\uparrow$ & Steps & MACs (G) & Para. (M) \\ \hline
Edge-SD-SR \cite{edgesdsr}                  &   \textcolor{blue}{24.10}    &  \textcolor{blue}{0.617}      &  \textcolor{red}{0.249}            &   25.37    & - &  \textcolor{green}{69.58}     &    1   &    -      &    \textcolor{blue}{169}     \\ \hline
AdcSR \cite{adcsr}                     & 23.74 & \textcolor{green}{0.602} & 0.285  & 25.52 & \textcolor{green}{4.36} & 68.00 & 1     & 496      & 456       \\ \hline
TinySR \cite{tinysr}                       &   -    &    0.572    &   \textcolor{green}{0.279}         &  \textcolor{blue}{22.94}    & \textcolor{blue}{4.15} & \textcolor{blue}{69.90}      &  1     &   427       &    341       \\ \hline
PocketSR \cite{pocketsr}     & \textcolor{green}{23.85} & 0.601 & 0.280  & \textcolor{green}{25.25} & 4.41 & 66.38 & 1    & \textcolor{red}{225}    & \textcolor{red}{146}      \\ \hline

 Nano-OSEDiff (Ours) &  \textcolor{red}{24.29}     &  \textcolor{red}{0.628}      &    0.296          &  27.46    & 4.92   &  66.41 &  1     &   \textcolor{green}{340}       &   448        \\ \hline
 Nano-S3Diff (Ours)  &  23.13     &  0.573      &       \textcolor{blue}{0.278}         &  \textcolor{red}{22.34}    &  \textcolor{red}{4.09}  &  \textcolor{red}{70.44} &    1   &   \textcolor{blue}{285}       &   \textcolor{green}{318}        \\ \hline
\end{tabular}
\end{table*}

 \begin{table*}[]
\centering
\caption{Quantitative comparison of different methods on synthetic CelebA-Test \cite{celebtest} dataset. The best, second-best and third-best results are highlighted in red, blue, and green colors, respectively. Refer to Supplemental for detailed analysis.}
\label{table:fr1}
\begin{tabular}{|l|l|l|l|l|l|l|l|l|l|}
\hline
Method              & LPIPS$\downarrow$ & DISTS$\downarrow$ & NIQE$\downarrow$ & MUSIQ$\uparrow$ & FID$\downarrow$    & LMD$\downarrow$  & Steps & MACs (G) & Para. (M) \\ \hline
PGDiff \cite{pgdiff}             & 0.386 & \textcolor{green}{0.195} & \textcolor{green}{4.001} & 69.56 & \textcolor{red}{44.62} & 7.310 & 1000  &     480998     &     \textcolor{blue}{176}      \\ \hline
DifFace \cite{difface}            & \textcolor{green}{0.346} & 0.212 & 4.638 & 66.74 & 49.80 & 5.475 & 250   &   46575       &     \textcolor{red}{175}      \\ \hline
DiffBIR \cite{diffbir}             & 0.374 & 0.234 & 6.280 & 75.63 & 71.77  & \textcolor{red}{5.104} & 50    &      24234    &   1717        \\ \hline
OSDFace \cite{osdface}            & \textcolor{red}{0.336} & \textcolor{red}{0.177} & \textcolor{red}{3.884} & \textcolor{green}{75.64} & \textcolor{blue}{45.41}  & 5.286 & 1     &   \textcolor{blue}{2465}     &   1887       \\ \hline
Nano-DiffBIR (Ours) & 0.382     & 0.212    & 6.3    & \textcolor{red}{76.52}     & 70.89     & \textcolor{green}{5.231}     & 50     &     \textcolor{green}{4734}     &     726      \\ \hline
Nano-OSDFace (Ours) & \textcolor{blue}{0.341}     & \textcolor{blue}{0.182}    & \textcolor{blue}{3.913}    & \textcolor{blue}{76.01}     & \textcolor{green}{45.92}     & \textcolor{blue}{5.172}     & 1     &    \textcolor{red}{479}      &    \textcolor{green}{415}       \\ \hline
\end{tabular}
\end{table*}

\section{Experiments}
\label{sec:experiments}

\subsection{Pareto Analysis and Model Selection}

We conduct two iterations of multi-objective search corresponding to the formulations described in Sec.~\ref{sec:method}.  
In the first iteration, we optimize taFID (our teacher-aligned FID metric; see Supplementary Material) jointly with on-device latency measured on a Qualcomm NPU using 8-bit weights and 16-bit activations.  
In the second iteration, we optimize taFID jointly with parameter count, which serves as a proxy for on-device memory usage.  
Fig.~\ref{fig:nas}e and Fig.~\ref{fig:nas}f show the resulting Pareto fronts for the latency--taFID and parameter--taFID objectives, respectively.  
In both plots, green circles denote Pareto-optimal points, while the remaining points are included for comparison.

Across the two searches, we obtain seven distinct Pareto-optimal architectures (Table~\ref{table:nas_arch}).  
Five models arise from the latency--accuracy front and three from the parameter--accuracy front, with one model appearing in both.  
Taken together, these architectures constitute the \emph{NanoSD family} and span a diverse set of trade-offs between generative fidelity and hardware efficiency.

Within this family, Model~5 achieves the lowest measured latency and represents the optimal choice when runtime efficiency is prioritized (\emph{NanoSD-Latency}).  
Model~7 attains the smallest parameter footprint and is preferred under strict memory constraints (\emph{NanoSD-Parameters}).  
Models~1 and~2 obtain the best taFID values overall with nearly identical accuracy.  
However, Model~1 incurs substantially higher latency, whereas Model~2 exhibits a more balanced accuracy--efficiency profile with only a marginal increase in parameter count (Table~\ref{table:nas_arch}).  
Because the latency difference between Models~1 and~2 is significantly larger than their parameter difference, we select Model~2 as the representative backbone for downstream evaluation and denote it as \emph{NanoSD-Prime}.  
The remaining models (1, 3, 4, and 6) occupy intermediate regions of the Pareto front and may be adapted to task-specific or resource-specific constraints (e.g., \texttt{NanoSD-S}, \texttt{NanoSD-XS}).

Fig.~\ref{fig:nas}e and Fig.~\ref{fig:nas}f also include two baselines: a manually designed hand-tuned variant of SD~1.5 and the prior-art Segmind TinySD \cite{tinysd} model.  
Both baselines lie far outside the Pareto frontier in terms of accuracy--latency and accuracy--parameter trade-offs.The hand-tuned model exhibits moderate latency but significantly worse taFID, indicating that manual structural simplifications fail to preserve the generative prior.  
TinySD achieves better accuracy but remains considerably slower and larger than all NanoSD variants.  

\begin{figure}
    \centering
    \includegraphics[width=1\linewidth]{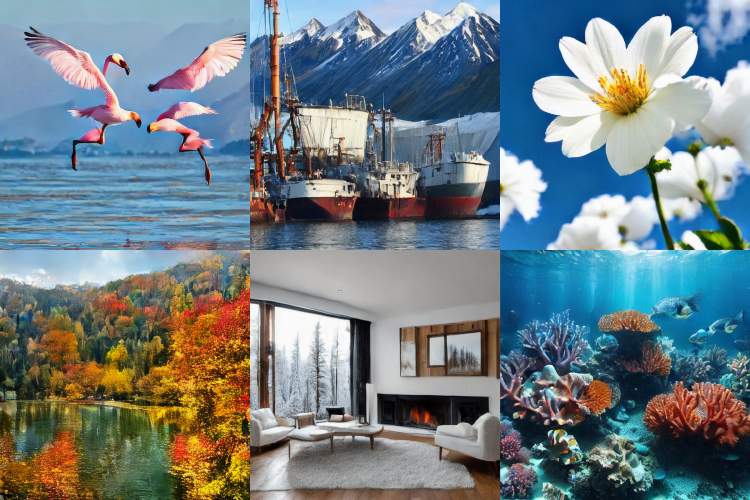}
    \caption{Generated images using edge efficient NanoSD.}
    \label{fig:nanosd}
\end{figure}

All seven NanoSD variants are subsequently fine-tuned end-to-end on image--text pairs derived from publicly available web-scale datasets (LAION-style \cite{laion}; see Supplementary Material).  
The final T2I FID scores for each model are reported in the Supplementary Tables.  
For the remainder of this paper, we adopt \emph{NanoSD-Prime} as our backbone \emph{NanoSD} (architecture shown in Fig.~\ref{fig:nas}h) and evaluate its generalization ability across a broad suite of low-level vision tasks. The generated images using the proposed NanoSD for T2I task are displayed in Fig.~\ref{fig:nanosd}.

\subsection{Experimental Settings}
\textbf{Datasets.} To test effectiveness of the proposed NanoSD, we utilize specific
datasets for each low-level task, SR: LSDIR \cite{lsdir} and the first 10K face images from
FFHQ \cite{ffhq}, face restoration: FFHQ \cite{ffhq}, deblurring: Gopro \cite{gopro}, dehazing: Reside \cite{reside}, deraining: merged train \cite{derain}, desnowing: Snow100K \cite{snow100k}, monecular depth estimation: Hypersim \cite{hypersim} and Virtual KITTI \cite{virtualkitti}. For testing, we evaluate on real-world benchmark datasets, SR: DIV2K-Val \cite{div2kval}, RealSR \cite{realsr} and DRealSR \cite{drealsr}, face restoration: Wider-Test \cite{codeformer}, LFWTest \cite{lfw}, and WebPhoto-Test \cite{webphoto}, deblurring: RealBlur-J \cite{realblurj}, dehazing: RTTS \cite{rtts}, deraining: real test \cite{realtest} and desnowing: realistic test \cite{realistic}, monocular depth estimation: NYU$_\text{{v2}}$ \cite{nyv2} and KITTI \cite{kitti}. \\  
\textbf{Evaluation Metrics.} To provide a comprehensive and holistic assessment on the performance of
different methods, we employ a range of full-reference and no-reference metrics. PSNR and SSIM \cite{ssim} (calculated on the Y channel in YCbCr space) are reference-based fidelity measures, while
LPIPS \cite{lpips}, DISTS \cite{dist} are reference-based perceptual quality measures. FID \cite{fid} evaluates the
distance of distributions between GT and restored images. NIQE \cite{niqe} and MUSIQ \cite{musiq}  are no-reference image quality measures. We apply two widely recognized metrics  Absolute Mean Relative Error (AbsRel) \cite{midasabsrel} and $\delta$1 accuracy \cite{dptdelta} for assessing quality of depth estimation.

\begin{figure*}
    \centering
    \includegraphics[width=1\linewidth]{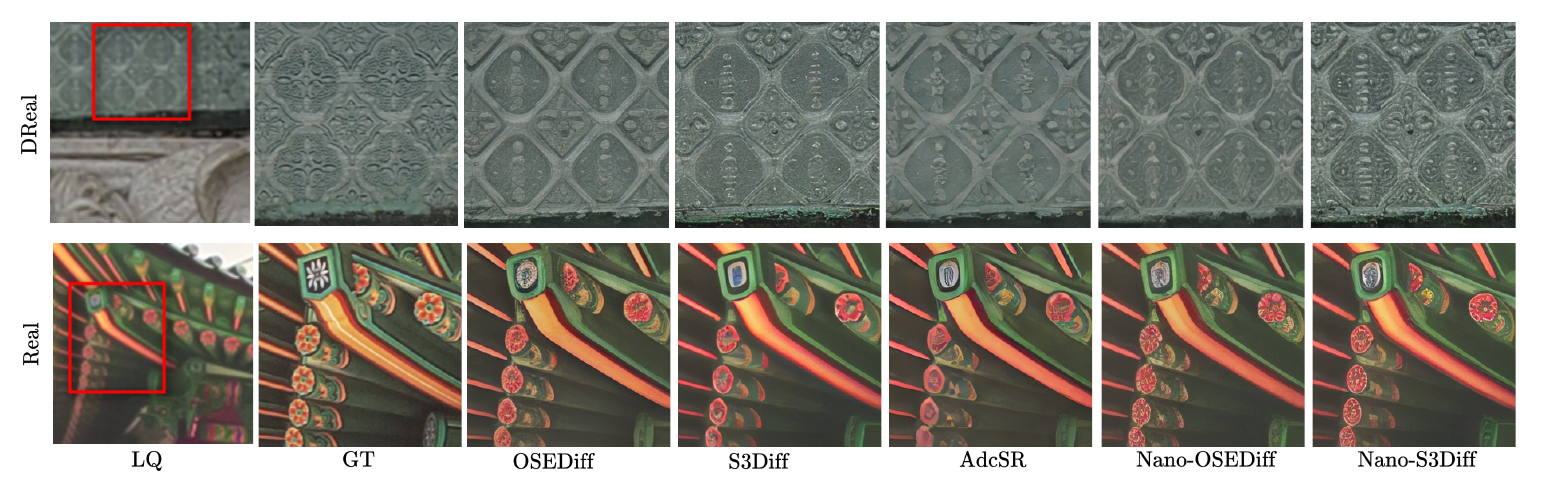}
    \caption{Qualitative comparisons of different SR methods. Please zoom in for a better view. Refer to Supplemental for more results.}
    \label{fig:sr}
\end{figure*}

\begin{figure*}[h]
    \centering
    \includegraphics[width=1\linewidth]{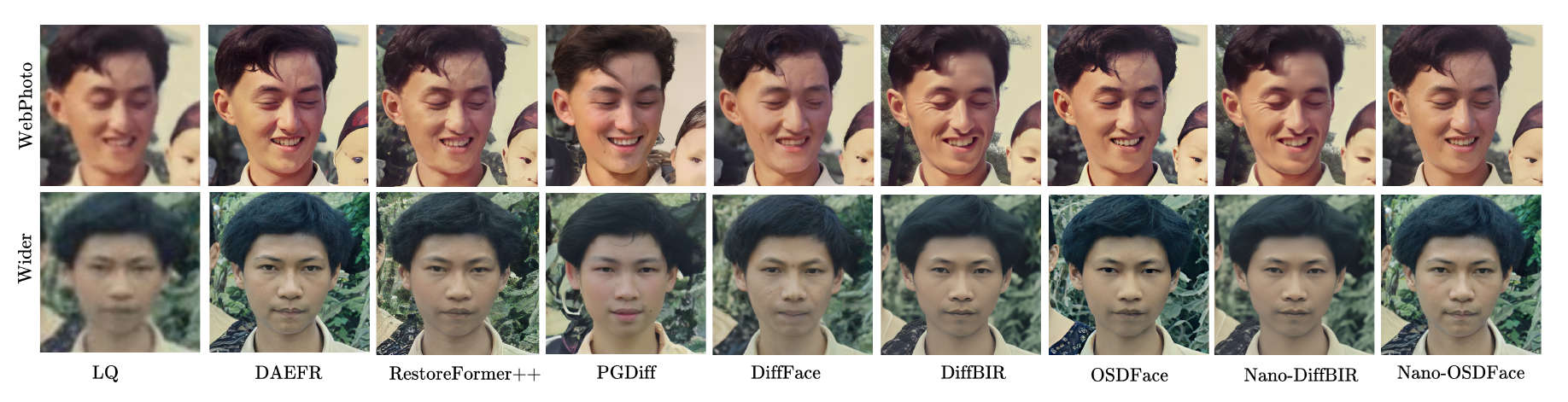}
    \caption{Visual comparisons of various face restoration methods. Please zoom in for a better view. Refer to Supplemental for more results.}
    \label{fig:fr}
\end{figure*}

\begin{table*}
\centering
\caption{Quantitative comparison of different methods on various low-level vision tasks. The best, second-best and third-best results are highlighted in red, blue, and green colors, respectively.}
\label{table:deblur}
\begin{tabular}{|l|ll|ll|ll|ll|l|l|l|}
\hline
\multirow{3}{*}{Method} & \multicolumn{2}{l|}{\multirow{2}{*}{\begin{tabular}[c]{@{}l@{}}Desnowing\\   Realstic \cite{realistic}\end{tabular}}} & \multicolumn{2}{l|}{\multirow{2}{*}{\begin{tabular}[c]{@{}l@{}}Dehazing\\    Reside \cite{reside}\end{tabular}}} & \multicolumn{2}{l|}{\multirow{2}{*}{\begin{tabular}[c]{@{}l@{}}Deblurring\\ RealBlur-J \cite{realblurj}\end{tabular}}} & \multicolumn{2}{l|}{\multirow{2}{*}{\begin{tabular}[c]{@{}l@{}}Deraining\\ Real Test \cite{realtest}\end{tabular}}} & \multirow{3}{*}{Steps} & \multirow{3}{*}{MACs(G)} & \multirow{3}{*}{Para.(M)} \\
                        & \multicolumn{2}{l|}{}                                                                                & \multicolumn{2}{l|}{}                                                                              & \multicolumn{2}{l|}{}                                                                                 & \multicolumn{2}{l|}{}                                                                               &                        &                           &                            \\ \cline{2-9}
                        & \multicolumn{1}{l|}{FID$\downarrow$}                                    & KID$\downarrow$                                    & \multicolumn{1}{l|}{FID$\downarrow$}                                   & KID$\downarrow$                                   & \multicolumn{1}{l|}{FID$\downarrow$}                                     & KID$\downarrow$                                    & \multicolumn{1}{l|}{FID$\downarrow$}                                    & KID$\downarrow$                                   &                        &                           &                            \\ \hline

SD \cite{sd}                     & \multicolumn{1}{l|}{35.24}                                  & 7.88                                   & \multicolumn{1}{l|}{48.89}                                 & 24.47                                 & \multicolumn{1}{l|}{59.21}                                   & 18.96                                  & \multicolumn{1}{l|}{51.78}                                  & 17.69                                 &      200                  &    79940                       &  1410                          \\ \hline
InstructP2P \cite{instructp2p}             & \multicolumn{1}{l|}{42.01}                                  & 8.54                                   & \multicolumn{1}{l|}{\textcolor{red}{33.48}}                                 & \textcolor{red}{12.76}                                 & \multicolumn{1}{l|}{57.38}                                   & 19.37                                  & \multicolumn{1}{l|}{54.12}                                  & 17.87                                 &        100                &    103381                       &            1027                \\ \hline
Null-Text \cite{nulltext}              & \multicolumn{1}{l|}{60.49}                                  & 16.38                                  & \multicolumn{1}{l|}{39.94}                                 & \textcolor{green}{14.88}                                 & \multicolumn{1}{l|}{60.38}                                   & 20.37                                  & \multicolumn{1}{l|}{\textcolor{green}{51.49}}                                  & 15.43                                 &         50               &    374137                       &                  1027          \\ \hline
ControlNet \cite{controlnet}             & \multicolumn{1}{l|}{\textcolor{blue}{34.36}}                                  & \textcolor{green}{5.70}                                   & \multicolumn{1}{l|}{37.02}                                 & 15.45                                 & \multicolumn{1}{l|}{\textcolor{blue}{52.30}}                                   & \textcolor{green}{17.19}                                  & \multicolumn{1}{l|}{52.55}                                  & \textcolor{green}{15.22}                                 &            50            &             \textcolor{blue}{24637}              &             1397               \\ \hline
Diff-Plugin \cite{diffplugin}            & \multicolumn{1}{l|}{\textcolor{red}{34.30}}                                  & \textcolor{red}{5.20}                                   & \multicolumn{1}{l|}{\textcolor{blue}{34.68}}                                 & \textcolor{blue}{14.38}                                 & \multicolumn{1}{l|}{\textcolor{red}{51.81}}                                   & \textcolor{red}{14.63}                                  & \multicolumn{1}{l|}{\textcolor{red}{50.55}}                                  & \textcolor{red}{13.84}                                 & 20                     &             \textcolor{green}{30400}              &             1256               \\ \hline
Nano-Diff-Plugin     & \multicolumn{1}{l|}{\textcolor{green}{34.83}}                                       &    \textcolor{blue}{5.63}                                    & \multicolumn{1}{l|}{\textcolor{green}{35.23}}                                      & 15.03                                       & \multicolumn{1}{l|}{\textcolor{green}{52.41}}                                        &   \textcolor{blue}{15.48}                                     & \multicolumn{1}{l|}{\textcolor{blue}{50.78}}                                       &      \textcolor{blue}{14.21}                                 & 20                       &             \textcolor{red}{17120}              &           712                 \\ \hline
\end{tabular}
\end{table*}

\subsection{Edge Efficient OSEDiff and S3Diff for Single Image Super-resolution}
 This paper aims to enable real-time super-resolution by effectively harnessing the rich generative priors inherent in pre-trained diffusion models. We validate the efficacy of the proposed foundation model for edge-efficient SR tasks by integrating NanoSD with prominent one-step latent diffusion frameworks, namely OSEDiff and S3Diff. 

Table \ref{table:sr1} presents quantitative comparisons of super-resolution methods across three datasets. Both proposed models exhibit second and third lowest MACs among evaluated approaches, enabling real-time processing and edge deployment. Nano-S3Diff achieves top NIQE and MUSIQ scores, along with second-best FID results, reflecting superior perceptual quality. Meanwhile, Nano-OSEDiff maintains competitive fidelity, yielding the superior PSNR and SSIM performance. Fig. \ref{fig:sr} presents qualitative comparisons between different super-resolution methods. Our proposed approaches demonstrate consistent performance in preserving image fidelity and reconstructing fine details. 

\subsection{Edge Efficient OSDFace for Face Restoration}
This section evaluates the applicability of our proposed model for face restoration tasks. To demonstrate its effectiveness, we integrate NanoSD into the OSDFace framework, which represents the current state-of-the-art one-step diffusion approach for face restoration.

Table \ref{table:fr1} presents quantitative comparisons on the CelebA-Test dataset. Our proposed Nano-OSDFace achieves comparable perceptual quality to OSDFace, attaining second-best scores in FID, NIQE, MUSIQ, and LMD which indicates alignment between generated and natural image distributions. The table further highlights Nano-OSDFace's efficiency, showing significant reduction in computational costs. Figure \ref{fig:fr} demonstrates that Nano-OSDFace effectively restores degraded facial inputs without deviating from the identity.

\begin{figure}
    \centering
    \includegraphics[width=1\linewidth]{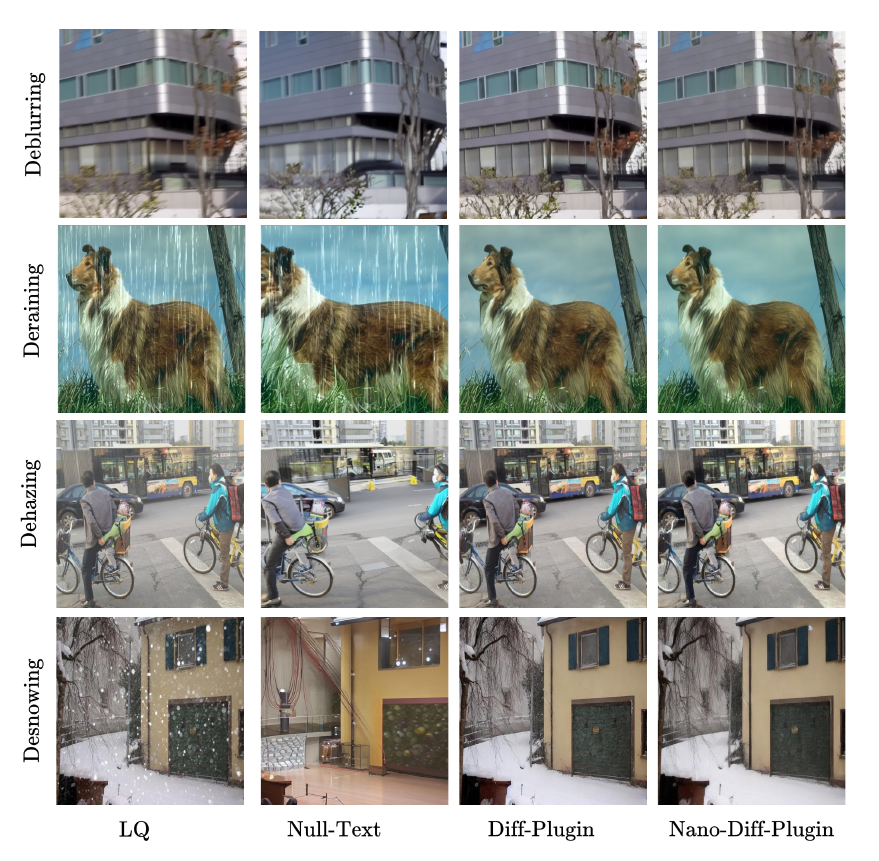}
    \caption{Qualitative comparisons of different methods for various restoration tasks. Please zoom in for a better view. Refer to Supplemental for more results.}
    \label{fig:diffplugin}
\end{figure}

\begin{table}[]
\centering
\caption{Quantitative comparison of the proposed Nano-Marigold with SOTA depth estimators on several zero-shot benchmarks. The best, second-best and third-best results are highlighted in red, blue, and green colors, respectively.}
\label{table:depth}
\begin{tabular}{|l|ll|ll|}
\hline
\multirow{2}{*}{Method} & \multicolumn{2}{l|}{NYU$_{V2}$ \cite{nyv2}}           & \multicolumn{2}{l|}{KITTI \cite{kitti}}          \\ \cline{2-5} 
                        & \multicolumn{1}{l|}{AbsRel $\downarrow$} & $\delta$1 $\uparrow$ & \multicolumn{1}{l|}{AbsRel $\downarrow$} & $\delta$1 $\uparrow$ \\ \hline
MiDAS \cite{midasabsrel}                  & \multicolumn{1}{l|}{11.1}   & 88.5  & \multicolumn{1}{l|}{23.6}   & 63.0  \\ \hline
LeRes \cite{lerecover}                  & \multicolumn{1}{l|}{9.0}    & 91.6  & \multicolumn{1}{l|}{14.9}   & 78.4  \\ \hline
Omnidata \cite{omnidata}               & \multicolumn{1}{l|}{7.4}    & 94.5  & \multicolumn{1}{l|}{14.9}   & 83.5  \\ \hline
HDN \cite{hdn}                     & \multicolumn{1}{l|}{\textcolor{blue}{6.9}}    & \textcolor{blue}{94.8}  & \multicolumn{1}{l|}{\textcolor{blue}{11.5}}   & \textcolor{green}{86.7}  \\ \hline
DPT \cite{dptdelta}                     & \multicolumn{1}{l|}{9.8}    & 90.3 & \multicolumn{1}{l|}{\textcolor{green}{10.0}}   & \textcolor{blue}{90.1}  \\ \hline
Marigold  \cite{marigold}              & \multicolumn{1}{l|}{\textcolor{red}{5.5}}    & \textcolor{red}{96.4}  & \multicolumn{1}{l|}{\textcolor{red}{9.9}}    & \textcolor{red}{91.6}  \\ \hline
Nano-Marigold        & \multicolumn{1}{l|}{\textcolor{green}{7.2}}       &  \textcolor{green}{94.6}     & \multicolumn{1}{l|}{11.8}       & 86.3      \\ \hline
\end{tabular}
\end{table}

\begin{figure}
    \centering
    \includegraphics[width=1\linewidth]{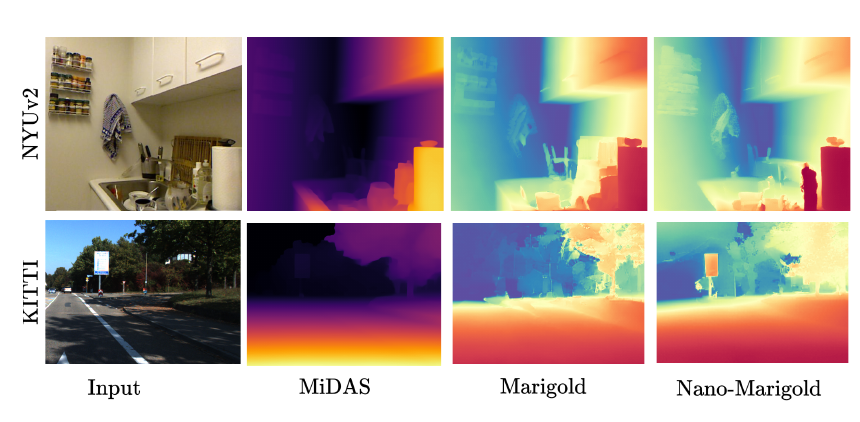}
    \caption{Qualitative comparison of various depth estimation methods across different datasets.}
    \label{fig:depth}
\end{figure}

\subsection{Accelerating Diff-Plugin and DiffBIR for Image Restoration with Generative Diffusion Prior}
The proposed NanoSD is trained on an extensive text-to-image dataset. To leverage its capabilities for diverse low-level vision tasks, we incorporate the model into the Diff-Plugin and DiffBIR frameworks. Both frameworks utilize a pre-trained latent diffusion backbone to adapt to multiple image restoration and enhancement tasks while preserving their inherent generative performance.

Table \ref{table:deblur} quantifies the performance of Nano-Diff-Plugin across multiple image restoration tasks. Our model achieves results comparable to state-of-the-art baselines while exhibiting substantial reductions in model size and inference time. Figure \ref{fig:diffplugin} illustrates the enhanced performance of Nano-Diff-Plugin across four challenging low-level vision tasks. 

Table \ref{table:fr1} demonstrates that the proposed Nano-DiffBIR achieves competitive performance in terms of IQA metrics and restores identity as indicated by lower LMD score. Visual comparisons in Figure \ref{fig:fr} reveal that the proposed approach effectively preserves facial identity while producing high-quality textures. 

\subsection{Accelerating Marigold: Repurposing NanoSD for Monocular Depth Estimation}
To harness the capabilities of a pretrained NanoSD model for monocular depth estimation, we incorporate it into Marigold—a latent diffusion framework tailored for depth prediction. The performance of Nano-Marigold in depth estimation is evaluated against alternative approaches in Table \ref{table:depth}. The results demonstrate that our method achieves competitive results compared to prior work while maintaining robust generalization across diverse real-world scenarios. For qualitative analysis, Figure \ref{fig:depth} presents a visual comparison. The results indicate that Nano-Marigold effectively captures both scene structure and fine-grained details with enhanced computational efficiency.

%% file: sec/conclusion.tex
\section{Conclusion}

We introduced NanoSD, an edge-efficient diffusion model obtained by restructuring the SD~1.5 U--Net into a hardware-aware search space, distilling stage-wise block variants, and selecting compact backbones through multi-objective optimization. The selected model, NanoSD preserves the generative behavior of SD~1.5 across a range of low-level vision tasks while substantially reducing computational cost on edge. 

%% file: sec/X_suppl.tex
\clearpage
\setcounter{page}{1}
\maketitlesupplementary

\section{Analysis of Generative Prior in NanoSD}

\begin{figure*}[ht]
    \centering
    \includegraphics[width=1\linewidth]{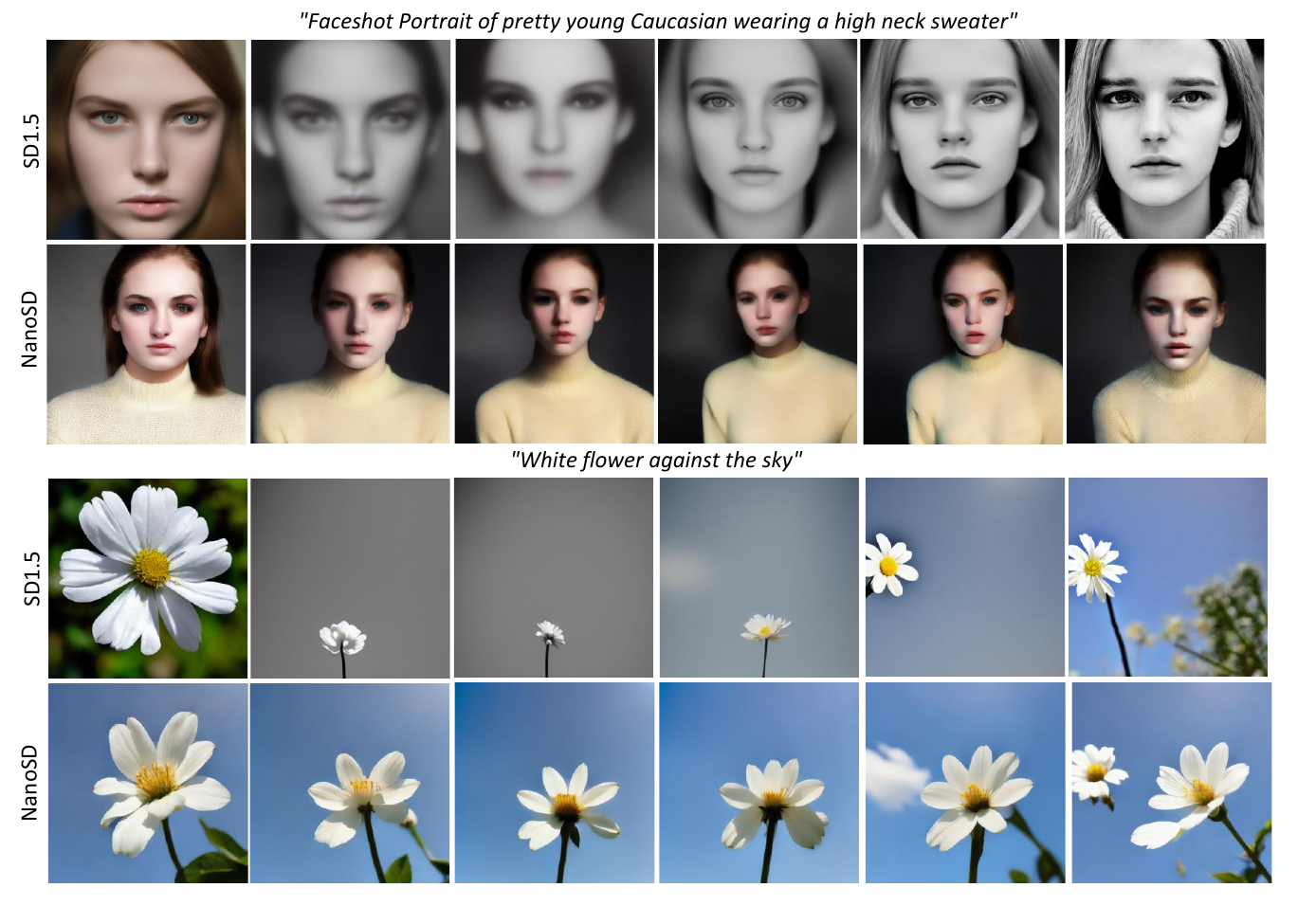}
    \caption{Latent-space interpolation between two random seeds for both SD~1.5 and NanoSD. The results demonstrate smooth transitions in the generated outputs along the interpolation trajectories for both models. This observation confirms that NanoSD preserves a well-structured latent manifold, maintaining consistency with the generative prior learned by SD~1.5.}
    \label{fig:latent_inter}
\end{figure*}

\begin{table*}[]
\centering
\caption{Perceptual distance measured using LPIPS and embedding-space similarity via CLIP embeddings between SD~1.5 outputs and other competitive models. Our results demonstrate that NanoSD maintains significantly closer proximity to SD~1.5 than the regression-based U-Net baseline. This quantitative evidence indicates robust preservation of SD~1.5's generative prior in the distilled NanoSD model.}
\label{tab:lpips_sim_comparison}
\begin{tabular}{|l|l|l|}
\hline
Model Pair & LPIPS $\downarrow$ & Embedding Cosine Similarity $\uparrow$ \\ \hline
SD~1.5 vs SD~1.5 (two random runs)  & 0.48 & 0.89 \\ \hline
NanoSD vs SD~1.5                   & 0.57 & 0.84\\ \hline
UNet-Baseline vs SD~1.5            & 1.92 & 0.41\\ \hline
\end{tabular}
\end{table*}

\begin{figure*}
    \centering
    \includegraphics[width=0.8\linewidth]{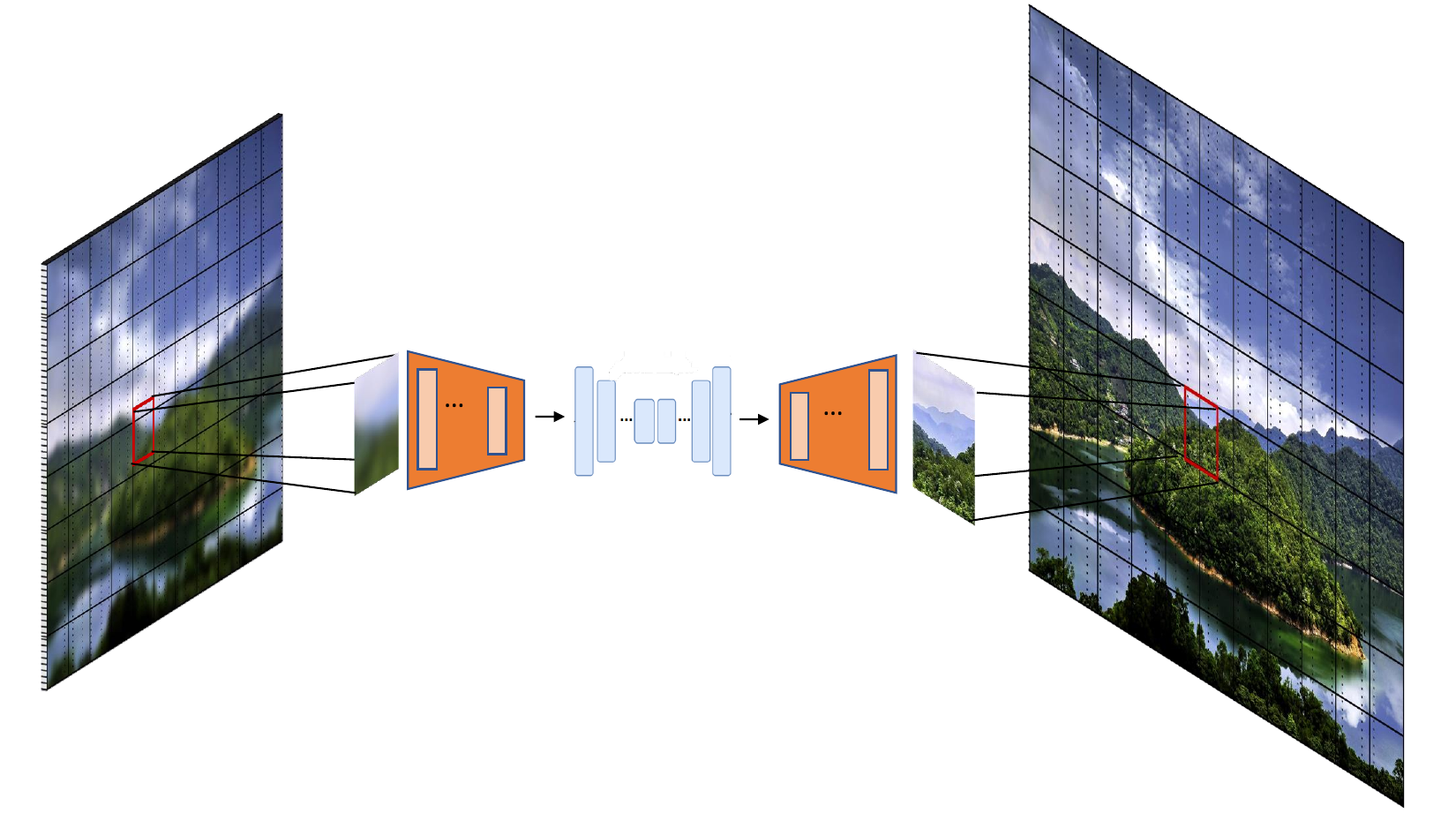}
    \caption{NanoSD employs a tiled inference strategy to enable high-resolution image restoration on edge computing platforms. The processing pipeline begins by partitioning a 1000$\times$750 input image into 128$\times$128 overlapping tiles (25$\%$ overlap), generating 88 total tiles. Each tile undergoes independent processing through the NanoSD model before being reassembled into a final 4K resolution output. This tile-based approach achieves two critical objectives: (1) maintaining perceptual consistency across the reconstructed image, and (2) meeting the computational constraints of mobile processors for real-time operation. The method effectively balances restoration quality with the practical requirements of edge deployment.}
    \label{fig:tiled_inference}
\end{figure*}

\begin{table*}[]
\centering
\caption{Latency profiling of 30 surrogate U--Net blocks on Qualcomm SM8750 NPU (Samsung S25 Ultra).
We report the measured per-block latency for all hardware-aware alternatives generated via network surgery. 
Although all variants preserve tensor shapes and remain accuracy-compatible after feature wise distillation, they exhibit substantial diversity in hardware cost, with several alternatives achieving 3-8$\times$ lower latency than the original SD~1.5 blocks. 
These measurements form the basis of the hardware-aware search space used in our Pareto optimization.}
\label{tab:surr_latency}
\begin{tabular}{|ll|ll|ll|}
\hline
\multicolumn{2}{|c|}{\textbf{Encoder 1}}                                                & \multicolumn{2}{c|}{\textbf{Encoder 2}}                                                & \multicolumn{2}{c|}{\textbf{Encoder 3}}                                                \\ \hline
\multicolumn{1}{|c|}{\textbf{Alternative}} & \multicolumn{1}{c|}{\textbf{Latency (ms)}} & \multicolumn{1}{c|}{\textbf{Alternative}} & \multicolumn{1}{c|}{\textbf{Latency (ms)}} & \multicolumn{1}{c|}{\textbf{Alternative}} & \multicolumn{1}{c|}{\textbf{Latency (ms)}} \\ \hline
\multicolumn{1}{|l|}{R}                    & 3                                          & \multicolumn{1}{l|}{R}                    & 1.79                                       & \multicolumn{1}{l|}{R}                    & 1.1                                        \\ \hline
\multicolumn{1}{|l|}{RA}                   & 12.5                                       & \multicolumn{1}{l|}{RA}                   & 3.7                                        & \multicolumn{1}{l|}{RA}                   & 1.46                                       \\ \hline
\multicolumn{1}{|l|}{RR}                   & 3.5                                        & \multicolumn{1}{l|}{RR}                   & 3                                          & \multicolumn{1}{l|}{RR}                   & 1.2                                        \\ \hline
\multicolumn{2}{|c|}{\textbf{Decoder 1}}                                                & \multicolumn{2}{c|}{\textbf{Decoder 2}}                                                & \multicolumn{2}{c|}{\textbf{Decoder 3}}                                                \\ \hline
\multicolumn{1}{|c|}{\textbf{Alternative}} & \multicolumn{1}{c|}{\textbf{Latency (ms)}} & \multicolumn{1}{c|}{\textbf{Alternative}} & \multicolumn{1}{c|}{\textbf{Latency (ms)}} & \multicolumn{1}{c|}{\textbf{Alternative}} & \multicolumn{1}{c|}{\textbf{Latency (ms)}} \\ \hline
\multicolumn{1}{|l|}{R}                    & 1.7                                        & \multicolumn{1}{l|}{R}                    & 2.07                                       & \multicolumn{1}{l|}{R}                    & 1.3                                        \\ \hline
\multicolumn{1}{|l|}{RA}                   & 16.3                                       & \multicolumn{1}{l|}{RA}                   & 5.8                                        & \multicolumn{1}{l|}{RA}                   & 3.4                                        \\ \hline
\multicolumn{1}{|l|}{RR}                   & 3.1                                        & \multicolumn{1}{l|}{RR}                   & 3.17                                       & \multicolumn{1}{l|}{RR}                   & 2.3                                        \\ \hline
\multicolumn{1}{|l|}{RRA}                  & 17.3                                       & \multicolumn{1}{l|}{RRA}                  & 6.8                                        & \multicolumn{1}{l|}{RRA}                  & 4.2                                        \\ \hline
\multicolumn{1}{|l|}{RAR}                  & 17.7                                       & \multicolumn{1}{l|}{RAR}                  & 7.5                                        & \multicolumn{1}{l|}{RAR}                  & 4.5                                        \\ \hline
\multicolumn{1}{|l|}{RRRA}                 & 19                                         & \multicolumn{1}{l|}{RRRA}                 & 8.8                                        & \multicolumn{1}{l|}{RRRA}                 & 5.5                                        \\ \hline
\multicolumn{1}{|l|}{RARR}                 & 19.1                                       & \multicolumn{1}{l|}{RARR}                 & 7.9                                        & \multicolumn{1}{l|}{RARR}                 & 5.2                                        \\ \hline
\multicolumn{1}{|l|}{RARA}                 & 39                                         & \multicolumn{1}{l|}{RARA}                 & 12.8                                       & \multicolumn{1}{l|}{RARA}                 & 5.5                                        \\ \hline
\end{tabular}
\end{table*}


\begin{table}[]
\centering
\caption{Cross-platform latency analysis of NanoSD family on the Apple A17 Pro Neural Engine (iOS). 
We evaluate all U--Net candidates from the NanoSD Pareto set—originally optimized using Qualcomm SM8750 latency data—on the Apple ANE without modification. 
The relative latency ordering and efficiency trends mirror those observed on SM8750, confirming that the proposed hardware-aware search produces architectures that generalize across accelerators. 
This consistency demonstrates the hardware-agnostic nature of our method and highlights the limitations of FLOP- or parameter-centric compression for predicting real-world edge performance.}
\label{tab:apple_latency}
\begin{tabular}{|c|cc|}
\hline
\multirow{2}{*}{\textbf{Model}} & \multicolumn{2}{c|}{\textbf{Latency (ms)}}              \\ \cline{2-3} 
                                & \multicolumn{1}{c|}{\textbf{Qualcomm}} & \textbf{Apple} \\ \hline
\textbf{TinySD}                 & \multicolumn{1}{c|}{74}                & 192            \\ \hline
\textbf{Hand-tuned}             & \multicolumn{1}{c|}{53}                & 133            \\ \hline
\textbf{NanoSD 1}               & \multicolumn{1}{c|}{41}                & 82             \\ \hline
\textbf{NanoSD 2}               & \multicolumn{1}{c|}{27}                & 38             \\ \hline
\textbf{NanoSD 3}               & \multicolumn{1}{c|}{24}                & 34             \\ \hline
\textbf{NanoSD 4}               & \multicolumn{1}{c|}{20}                & 31             \\ \hline
\textbf{NanoSD 5}               & \multicolumn{1}{c|}{12}                & 20             \\ \hline
\textbf{NanoSD 6}               & \multicolumn{1}{c|}{28}                & 68             \\ \hline
\textbf{NanoSD 7}               & \multicolumn{1}{c|}{27}                & 66             \\ \hline
\end{tabular}
\end{table}

A central claim of this work is that NanoSD retains the generative prior inherited from the original SD~1.5 backbone, even after undergoing distillation and downstream task adaptation. To validate this claim, we perform comprehensive qualitative and quantitative analyses, comparing NanoSD with both SD~1.5 and a computationally equivalent UNet-based restoration baseline.

\subsection{Exploring the Latent Space Manifold}
To evaluate whether NanoSD preserves the structured latent manifold inherent to SD~1.5, we perform latent-space interpolation between two Gaussian noise samples. As demonstrated in Fig.~\ref{fig:latent_inter}, both SD~1.5 and NanoSD exhibit smooth and semantically consistent transitions along the interpolation path, maintaining scene layout and object semantics. This behavior suggests that NanoSD successfully retains the high-level generative structure of the original diffusion model.

\subsection{Exploring Embedding-Based Similarity and Perceptual Consistency}
To quantitatively assess prior preservation, we measure embedding-based similarities using CLIP encoders. If NanoSD successfully retains the generative prior of SD~1.5, its outputs should maintain proximity to the distribution generated by the original SD~1.5 model. As evidenced in Table~\ref{tab:lpips_sim_comparison}, NanoSD demonstrates significantly higher similarity to SD~1.5 compared to the U-Net baseline, while approaching the self-consistency of SD~1.5 evaluated across independent sampling runs. These results indicate that NanoSD effectively preserves the generative manifold of the original SD~1.5 model.

We further assess perceptual fidelity using the Learned Perceptual Image Patch Similarity (LPIPS)~\cite{lpips} metric for image super-resolution. As shown in Table~\ref{tab:lpips_sim_comparison}, NanoSD achieves a substantially lower perceptual distance to SD~1.5 compared to the UNet baseline. Crucially, the LPIPS difference between NanoSD and SD~1.5 is only marginally larger than the variation observed between different SD~1.5 outputs generated with distinct random seeds. This narrow margin demonstrates that NanoSD maintains strong alignment with the teacher model's output distribution despite the distillation process.

\section{Real-time Image Restoration using NanoSD}

\subsection{Mobile ISP for Image Restoration}

We address the significant challenge of developing a foundation model capable of preserving generative priors while operating efficiently on resource-constrained edge devices. Deploying such models on mobile platforms presents multiple practical obstacles, including large model sizes and high inference latency. These challenges are particularly pronounced for restoration tasks, which typically require processing high-resolution images through multiple model runs. For instance, modern smartphones often capture images at 4K$\times$3K resolution, whereas diffusion-based models like StableSR typically generate 512$\times$512 patches per forward pass. This discrepancy necessitates tile-based processing, where a 1000$\times$750 low-quality input image is divided into 128$\times$128 patches with 25$\%$ overlap, resulting in 88 patches that must be processed sequentially as shown in Fig~\ref{fig:tiled_inference}. Each tile undergoes one or more diffusion steps, substantially increasing computational overhead and runtime.

\subsection{Limitations of SD~1.5 for Edge Deployment}

While diffusion models such as SD~1.5 offer strong generative priors, their computational and memory demands make them fundamentally unsuitable for resource-constrained devices. SD~1.5 contains approximately 829 million parameters, corresponding to about 3.3~GB of raw model weights in Full precision. Even with INT8 quantization, the footprint remains roughly 0.8~GB, exceeding the memory budgets of mobile NPUs, which must also accommodate activation buffers, intermediate tensors, runtime libraries, and system services. This alone renders SD~1.5 impractical for on-device deployment.

In practice, the situation is even more restrictive. Attempts to deploy SD~1.5 on commercial mobile chipsets failed to generate executable binaries for both Qualcomm NPUs (Android) and the Apple Neural Engine (iOS) when using FP32, INT16, or INT8 precision. Even aggressive INT4 quantization—despite its substantial accuracy degradation—yielded a model with an average latency of 116~ms per tile on the Qualcomm Snapdragon SM8750 NPU.

Recalling the tile-based pipeline described previously, a $1000 \times 750$ input image requires 88 tiles at $128 \times 128$ resolution. Thus, the end-to-end runtime of SD~1.5 becomes 10.2 seconds (~88 tiles x 116 ms/tile).
This latency corresponds to a \emph{single} diffusion step, whereas restoration pipelines typically require multiple steps or sequential subnetworks (e.g., face refinement or iterative enhancement). Such runtimes are incompatible with real-time or interactive workloads. Together, these memory, compilation, and runtime constraints demonstrate why SD~1.5 is unsuitable for deployment on modern mobile devices and motivate the need for an edge-aware redesign.

\subsection{Hardware-Aware Network Surgery}

To bridge this gap, we perform a principled hardware-aware decomposition of the SD~1.5 U--Net. Rather than applying uniform pruning or parameter-centric compression, we restructure the network at the block level through \emph{network surgery}. Specifically, we identify alternative block configurations—such as residual-only, reduced-attention, or hybrid arrangements—that preserve the input--output tensor shapes of the original SD~1.5 blocks. These alternatives are then individually profiled on the Qualcomm SM8750 NPU, enabling us to measure the true hardware latency of each candidate block.

This profiling reveals substantial variation in runtime among shape-compatible alternatives, with several variants offering 3--8$\times$ lower latency despite comparable parameter counts. Table~\ref{tab:surr_latency} lists all 30 surrogate blocks and their measured latencies on the SM8750 NPU in full precision. Such profiling enables us to embed hardware cost directly into the design space, avoiding reliance on FLOPs or parameter count, which correlate poorly with real-world latency due to tensor-movement and memory-access bottlenecks.

While hardware profiling identifies efficient block candidates, directly substituting them into the U--Net would disrupt SD~1.5's generative prior. To preserve fidelity, we apply \emph{Feature-wise Generative Distillation} (FwGD) to every candidate block: each surrogate is trained to match the output distribution of its corresponding SD~1.5 teacher block under identical inputs. The structural compatibility between the base block and each hardware-aware alternative ensures the path for knowledge transfer to compact, local modules, enabling massively parallel and computationally lightweight training.

FwGD provides two essential benefits:
\begin{itemize}
    \item \textbf{Accuracy preservation.} Each block independently inherits SD~1.5’s generative behavior, allowing arbitrary combinations of surrogates to form valid U--Nets without catastrophic feature drift.
    \item \textbf{Search-space validity.} By distilling all block variants, the search space contains only architectures that are simultaneously hardware-efficient and accuracy-sufficient.
\end{itemize}

Together, hardware-aware network surgery and FwGD yield a search space that is structurally valid, hardware-aligned, and generative-prior-preserving—properties crucial for downstream optimization.

\subsection{Why Hardware Awareness Is Essential?}

If the search space were constructed solely using parameter reduction, FLOPs minimization, or uniform pruning, the resulting architectures would not necessarily yield improvements in \emph{true} edge latency. In hierarchical U--Nets, inner blocks operate at low spatial resolutions, so reducing their parameters has minimal effect on runtime. Conversely, even small modifications to early, high-resolution stages significantly reduce latency due to the cost of propagating large activation maps. This mismatch between compute-centric metrics and latency-critical behavior explains why SD~1.5 cannot simply be ``compressed'' into an edge model.

By profiling block alternatives directly on NPU hardware, our method ensures that latency-critical regions receive appropriate structural simplifications. Combined with FwGD, this yields a hardware-aware and accuracy-aligned search space from which Bayesian optimization extracts Pareto-optimal U--Nets, ultimately producing the NanoSD architecture. This synergy between hardware profiling, network surgery, and block-level distillation fundamentally differentiates our method from FLOP-centric or parameter-centric reduction techniques.

\subsection{Generic Nature of the Proposal}

To assess the generality of our approach, we replicated the hardware profiling procedure on a second, architecturally distinct platform: the Apple A17 Pro Neural Engine (ANE) on iOS. Although the network surgery and Pareto optimization were performed using latency measurements from the Qualcomm SM8750 NPU on the Samsung S25 Ultra, we evaluated the resulting NanoSD family on the ANE without modification. As shown in Table~\ref{tab:apple_latency}, the latency trends on the ANE closely mirror those observed on the SM8750: NanoSD variants consistently outperform the SD~1.5 baseline and manually designed alternatives despite differences in compiler stacks, operator scheduling, and memory hierarchies across the two platforms. This consistency demonstrates that the proposed method is inherently hardware-agnostic: rather than tailoring a model to a single device, it constructs a search space aware of device characteristics while remaining compatible with any backend NPU. These results reinforce that FLOPs and parameter count alone are insufficient predictors of on-device performance, and that explicit hardware-aware decomposition is essential for designing diffusion models suitable for diverse edge environments.

\subsection{Edge efficiency of NanoSD }

We conduct a systematic evaluation of NanoSD's computational efficiency to assess its viability for edge deployment in high-resolution image restoration applications. Our quantitative analysis reveals that NanoSD achieves inference latencies compatible with real-time operational requirements on mobile platforms. Specifically, the model processes a 128$\times$128 input tile to produce a 512$\times$512 output in approximately 20 ms when running on mobile neural processing units (NPUs). For processing 4000$\times$3000 image, the total end-to-end latency measures of about 1.8 seconds on equivalent hardware, accounting for all 88 required tile evaluations. These empirical results provide compelling evidence that diffusion-based restoration can be effectively implemented on-device, satisfying the stringent computational constraints of mobile platforms while eliminating the need for cloud-based processing.

\section{Bayesian Optimization with EHVI}

We cast the selection of an efficient U--Net backbone from the discrete hardware-aware search space as a multi-objective black-box optimization problem. Let $\mathcal{Z}$ denote the discrete set of feasible architectures induced by stage-wise block choices. For any architecture $\mathbf{z}\in\mathcal{Z}$ we evaluate two primary objectives:
\begin{equation}
f_{\mathrm{FID}}(\mathbf{z}) 
= \mathrm{FID}\!\left( \hat{X}(\mathbf{z}), X_{\mathrm{SD1.5}} \right),
\label{eq:fid}
\end{equation}

\begin{equation}
f_{\mathrm{param}}(\mathbf{z}), \qquad 
f_{\mathrm{latency}}(\mathbf{z}).
\label{eq:efficiency_objs}
\end{equation}

where $f_{FID}$, the first objective ($f_{1}$), is the teacher-aligned FID measured against SD~1.5 outputs, the second objective ($f_{2}$) is either $f_{latency}$, the on-device latency or $f_{param}$, the parameter count, depending on the search iteration.

Direct optimization over $\mathcal{Z}$ is expensive because each evaluation of $f_1$ requires many diffusion samples and FID computation. We therefore employ Bayesian Optimization (BO) with Expected Hypervolume Improvement (EHVI) ~\cite{snoek2012practical} as the acquisition function to efficiently explore $\mathcal{Z}$. To handle the discrete nature of $\mathcal{Z}$ we adopt a two-step relaxed-and-project approach: (1) represent architecture choices by a continuous proxy $\mathbf{x}\in[0,1]^d$ (here $d$ is the number of design dimensions), (2) map $\mathbf{x}$ to the nearest feasible discrete architecture via a deterministic projection $\phi:\,[0,1]^d \to \mathcal{Z}$.

The BO loop proceeds as follows. We fit independent Gaussian Process (GP) surrogates to the observed values of each objective over previously evaluated architectures. EHVI is computed with respect to the current Pareto set and used to propose the next continuous candidate $\mathbf{x}^*$. The projection $\mathbf{z}^*=\phi(\mathbf{x}^*)$ yields a discrete architecture which is assembled from the distilled surrogate blocks and then evaluated to obtain $(f_1(\mathbf{z}^*), f_2(\mathbf{z}^*))$. The GP models are updated and the Pareto set is expanded accordingly. Iteration continues until a predefined budget of evaluations is exhausted, resulting the final Pareto set as shown in Figure 2 in the main paper.

\section{VAE Distillation}

Our objective is to obtain a fully edge-optimized latent diffusion model by compressing all major components of Stable Diffusion~1.5-the U-Net, VAE encoder, and VAE decoder-while preserving compatibility with the teacher model’s generative behavior. Starting from a pretrained SD~1.5 pipeline, we first distill the U-Net, then distill the VAE encoder and decoder with the U-Net frozen, and finally refine the assembled pipeline for end-to-end consistency.

Table~\ref{tab:vae_architecture_comparison} provides an overview and includes the distilled architectures of both the encoder and decoder. The proposed student VAE is highly parameter- and latency-efficient: its encoder and decoder contain only $\sim$2 M and $\sim$1.3 M parameters (10 ms and 8 ms FP16 latency, respectively), compared to the much larger SD,1.5 VAE. This reduction is achieved by using fixed-width (64-channel) Tiny ResNet blocks and shallow lightweight up/downsampling stages instead of the widening 64→128→256→512 ResNet blocks used in the teacher.

\begin{table*}[t]
\centering
\caption{Architectural comparison of the Teacher VAE (SD\,1.5) and the proposed Student VAE. Both models are shown using a unified notation with identical spatial dimensions. Only architectural differences are listed. ResNetBlockTiny refers to a lightweight residual block that preserves
the Conv–Norm–Activation–Conv pattern of a standard ResNet block but replaces the
full $3\times3$ convolutions with parameter–efficient variants (e.g. depthwise–separable convs) operating at a fixed 64-channel width.
This makes each block an order of magnitude smaller than the full ResNet blocks
in the SD\,1.5 encoder.}

\renewcommand{\arraystretch}{1.32}
\begin{tabular}{|p{0.46\linewidth}|p{0.46\linewidth}|}
\hline
\multicolumn{1}{|c|}{\textbf{Teacher VAE (SD\,1.5)}} &
\multicolumn{1}{c|}{\textbf{Student VAE (Proposed)}} \\
\hline
\multicolumn{2}{|c|}{\textbf{Encoder Architecture (Input: $3\times H\times W$)}} \\
\hline

\textbf{Stem:}  
Conv(3→64), ResNetBlock ×2.  
Resolution preserved.  
\vspace{4pt}

\textbf{Down Block 1:}  
ResNetBlock ×2 @ 64 ch, stride-2 Conv → $H/2 \times W/2$.  
\vspace{4pt}

\textbf{Down Block 2:}  
ResNetBlock ×2 @ 128 ch, stride-2 Conv → $H/4 \times W/4$.  
\vspace{4pt}

\textbf{Down Block 3:}  
ResNetBlock ×2 @ 256 ch, stride-2 Conv → $H/8 \times W/8$.  
\vspace{4pt}

\textbf{Middle/Bottleneck:}  
ResNetBlock ×2 @ 512 ch.  
\vspace{4pt}

\textbf{Output projection:}  
Conv→8 ch.  
&
\textbf{Stem:}  
Conv(3→64), \textbf{ResNetBlockTiny ×1}.  
Resolution preserved.  
\vspace{4pt}

\textbf{Down Block 1:}  
\textbf{ResNetBlockTiny ×3} @ 64 ch, stride-2 Conv → $H/2 \times W/2$.  
\vspace{4pt}

\textbf{Down Block 2:}  
\textbf{ResNetBlockTiny ×3} @ 64 ch, stride-2 Conv → $H/4 \times W/4$.  
\vspace{4pt}

\textbf{Down Block 3:}  
\textbf{ResNetBlockTiny ×3} @ 64 ch, stride-2 Conv → $H/8 \times W/8$.  
\vspace{4pt}

\textbf{Middle/Bottleneck:}  
(no dedicated bottleneck; final stage already at 64 ch).  
\vspace{4pt}

\textbf{Output projection:}  
Conv→4 ch. \\

\hline
\multicolumn{2}{|c|}{\textbf{Decoder Architecture (Input: $z \in \mathbb{R}^{4\times (H/8)\times(W/8)}$)}} \\
\hline

\textbf{Stem:}  
Conv(4→512).  
\vspace{4pt}

\textbf{Middle/Bottleneck:}  
ResNetBlock ×2 @ 512 ch.  
\vspace{4pt}

\textbf{Up Block 1:}  
Upsample×2, ResNetBlock ×2 (512→256) → $H/4 \times W/4$.  
\vspace{4pt}

\textbf{Up Block 2:}  
Upsample×2, ResNetBlock ×2 (256→128) → $H/2 \times W/2$.  
\vspace{4pt}

\textbf{Up Block 3:}  
Upsample×2, ResNetBlock ×2 (128→64) → $H \times W$.  
\vspace{4pt}

\textbf{Final projection:}  
Conv(64→3).  
&
\textbf{Stem:}  
Conv(4→64).  
\vspace{4pt}

\textbf{Middle/Bottleneck:}  
ResNetBlock ×1 @ 64 ch.  
\vspace{4pt}

\textbf{Up Block 1:}  
ResNetBlock ×2 @ 64 ch, Upsample×2 → $H/4 \times W/4$.  
\vspace{4pt}

\textbf{Up Block 2:}  
ResNetBlock (→128 ch), Upsample×2 → $H/2 \times W/2$.  
\vspace{4pt}

\textbf{Up Block 3:}  
ResNetBlock @ 128 ch, Upsample×2 → $H \times W$.  
\vspace{4pt}

\textbf{Final projection:}  
Conv(128→3). \\

\hline
\end{tabular}
\label{tab:vae_architecture_comparison}
\end{table*}

\subsection{Preliminaries}

Let $E_t$, $D_t$, and $U_t$ denote the teacher VAE encoder, VAE decoder, and diffusion U-Net. The encoder outputs posterior parameters
\begin{equation}
(\mu_t(x), \log\sigma_t^2(x)) = E_t(x).
\label{eq:teacher_posterior}
\end{equation}
Latent sampling follows
\begin{equation}
z_t = \mu_t(x) + \sigma_t(x)\odot\epsilon, \qquad \epsilon \sim \mathcal{N}(0,I).
\label{eq:reparam}
\end{equation}
Scaled latents are fed into the U-Net:
\begin{equation}
\tilde{z}_t = \alpha z_t,\qquad \alpha = 0.18215.
\label{eq:latent_scale}
\end{equation}
The U-Net predicts noise conditioned on time step $t$ and text embedding $c$:
\begin{equation}
\epsilon_t = U_t(\tilde{z}_t, t, c).
\label{eq:unet_teacher_predict}
\end{equation}

Our goal is to obtain student modules $E_s$, $D_s$, and $U_s$ such that the resulting pipeline remains distributionally aligned with the teacher.

\subsection{Stage I: U-Net Distillation}

The student U-Net $U_s$ is trained via the proposed formulation (see Figure 2 in main paper) for edge efficiency, resulting in the NanoSD U-Net. After this stage, $U_s$ is frozen for the remainder of training.

\subsection{Stage II: Encoder Distillation}

The student encoder outputs
\begin{equation}
(\mu_s(x), \log\sigma_s^2(x)) = E_s(x),
\label{eq:student_encoder}
\end{equation}
and is trained to match the teacher encoder via
\begin{equation}
\mathcal{L}_{\text{latent}}
=
\|\mu_t(x) - \mu_s(x)\|_2^2
+
\|\sigma_t(x) - \sigma_s(x)\|_2^2.
\label{eq:latent_loss}
\end{equation}

A weak KL regularization term is included:
\begin{equation}
\mathcal{L}_{\text{KL}}
=
D_{\mathrm{KL}}(q_s(z\mid x)\|\mathcal{N}(0,I)).
\label{eq:kl_loss}
\end{equation}

The encoder objective is:
\begin{equation}
\mathcal{L}_{E}
=
\mathcal{L}_{\text{latent}}
+
\beta\,\mathcal{L}_{\text{KL}},
\qquad
\beta\in[10^{-6},10^{-4}].
\label{eq:encoder_objective}
\end{equation}

\subsection{Stage III: Decoder Distillation}

Given teacher latents $z_t$, the teacher and student reconstructions are:
\begin{equation}
\hat{x}_t = D_t(z_t), \qquad \hat{x}_s = D_s(z_t).
\label{eq:decoder_outputs}
\end{equation}

Teacher-guided reconstruction is enforced via
\begin{equation}
\mathcal{L}_{\text{recon}}
=
\|\hat{x}_t - \hat{x}_s\|_1.
\label{eq:recon_loss}
\end{equation}

A perceptual loss aligns higher-level features:
\begin{equation}
\mathcal{L}_{\text{perc}}
=
\|\phi(\hat{x}_t) - \phi(\hat{x}_s)\|_2^2.
\label{eq:perc_loss}
\end{equation}

To ensure decoder robustness to student latents $z_s$, we include
\begin{equation}
\mathcal{L}_{\text{AE}}
=
\|x - D_s(z_s)\|_1.
\label{eq:ae_loss}
\end{equation}

The decoder objective is thus
\begin{equation}
\mathcal{L}_{D}
=
\mathcal{L}_{\text{recon}}
+
\lambda_{\text{perc}}\mathcal{L}_{\text{perc}}
+
\lambda_{\text{AE}}\mathcal{L}_{\text{AE}}.
\label{eq:decoder_objective}
\end{equation}

\subsection{Stage IV: Joint VAE Refinement}

Once $E_s$ and $D_s$ are individually trained, we jointly refine them using
\begin{equation}
\mathcal{L}_{\text{VAE}}
=
\lambda_1\|x - D_s(E_s(x))\|_1
+
\lambda_2\mathcal{L}_{\text{perc}}
+
\lambda_3 D_{\mathrm{KL}}(q_s\|\mathcal{N}(0,I)).
\label{eq:vae_joint}
\end{equation}

\subsection{Stage V: End-to-End Diffusion Alignment}

To compensate for latent shift introduced by VAE compression, we lightly fine-tune $U_s$ while keeping $E_s$ and $D_s$ frozen:
\begin{equation}
\mathcal{L}_{\text{align}}
=
\|
U_s(\alpha z_s + \sigma_t\epsilon, t, c)
-
U_t(\alpha z_t + \sigma_t\epsilon, t, c)
\|_2^2.
\label{eq:align_loss}
\end{equation}

This multi-stage procedure yields a compact, edge-efficient latent diffusion model that preserves the generative behavior of SD~1.5 while enabling practical on-device deployment.

\section{Implementation Details}

\paragraph{Hardware for Latency Profiling.}
All hardware-aware latency measurements were performed on a Qualcomm Snapdragon SM8750 NPU deployed on a Samsung Galaxy S25 Ultra. Each of the 30 block alternatives was compiled using Qualcomm SNPE with INT8 weights and FP16 activations, and latency was averaged over 256 runs after discarding the first 20 warm-up iterations. 
To assess generality across heterogeneous hardware, the final NanoSD U-Net family was also profiled on the Apple A17 Pro Neural Engine using CoreML Tools on iOS~17, where a consistent latency ranking was observed (Table~\ref{tab:apple_latency}).

\paragraph{Dataset for Distillation.}
Feature-wise Generative Distillation, VAE distillation, and end-to-end alignment were all conducted using a representative subset of approximately 1 million image-text pairs sampled from a LAION-style corpus. 
This subset was stratified across prompt categories and image types to ensure coverage equivalent to the teacher model's operating distribution. A held-out validation subset of 50,000 samples was used for early stopping and monitoring drift.

\paragraph{Feature-wise Generative Distillation of 30 Surrogates.}
Each of the 30 U-Net surrogate blocks was distilled independently against the corresponding SD~1.5 teacher block.  
We used AdamW (learning rate $1\times 10^{-4}$, weight decay $0.01$), batch size 128, cosine decay schedule with 2k warmup steps, and 15k--40k training iterations depending on block complexity.  
The total compute budget for all 30 surrogates was approximately 360 A100 GPU-hours.  
Parallel training across multiple GPUs allowed all surrogates to be distilled simultaneously.

\paragraph{Bayesian Optimization Configuration.}
Bayesian optimization was applied separately for the two bi-objective settings:  
(i) taFID vs. edge latency, and  
(ii) taFID vs. parameter count.  
We used a 6-dimensional continuous relaxation of the discrete search space, projected to feasible architectures via a deterministic thresholding map. Independent Gaussian Processes with Matérn-5/2 kernels were fitted to both objectives.  
The acquisition function was Expected Hypervolume Improvement (EHVI).  
We used 15 random initial samples and 120 BO iterations per run, and performed 5 independent runs with different seeds.  
The total compute cost for BO evaluation-dominated by taFID computation-was approximately 100 A100 GPU-hours.

\paragraph{VAE Encoder and Decoder Distillation.}
After selecting the NanoSD U-Net, we distilled the VAE encoder and decoder using the same 1M-sample dataset.  
The encoder was trained with AdamW (learning rate $5\times10^{-5}$, batch size 64) for 40k iterations, minimizing latent-matching and KL regularization losses with $\beta \in [10^{-6}, 10^{-4}]$.  
The decoder was trained for 80k--120k iterations with AdamW (learning rate $1\times10^{-4}$), using reconstruction, perceptual, and autoencoding consistency losses.  
Combined encoder+decoder distillation required approximately ~220 A100 GPU-hours.

\paragraph{End-to-End Diffusion Alignment.}
To compensate for the mild latent shift introduced by VAE compression, we performed a lightweight end-to-end noise-prediction alignment stage.  
The NanoSD U-Net was fine-tuned with a small learning rate ($1\times10^{-6}$) for 20k--40k steps, with the VAE frozen.  
We used batch size 32, random timesteps, and kept the text encoder frozen.  
This stage required an additional ~70 A100 GPU-hours.  

\paragraph{Total Compute.}
Across all stages (FwGD, BO evaluations, VAE distillation, and T2I end-to-end alignment), the complete NanoSD pipeline required approximately 750 A100 GPU-hours-substantially lower than training or distilling a full SD~1.5 model, while enabling hardware-aware optimization tailored to edge devices.

\section{Experimental Results}
We present a detailed experimental and ablation analysis of various low-level vision tasks using the proposed NanoSD.

\begin{figure}[ht]
    \centering
    \includegraphics[width=1\linewidth]{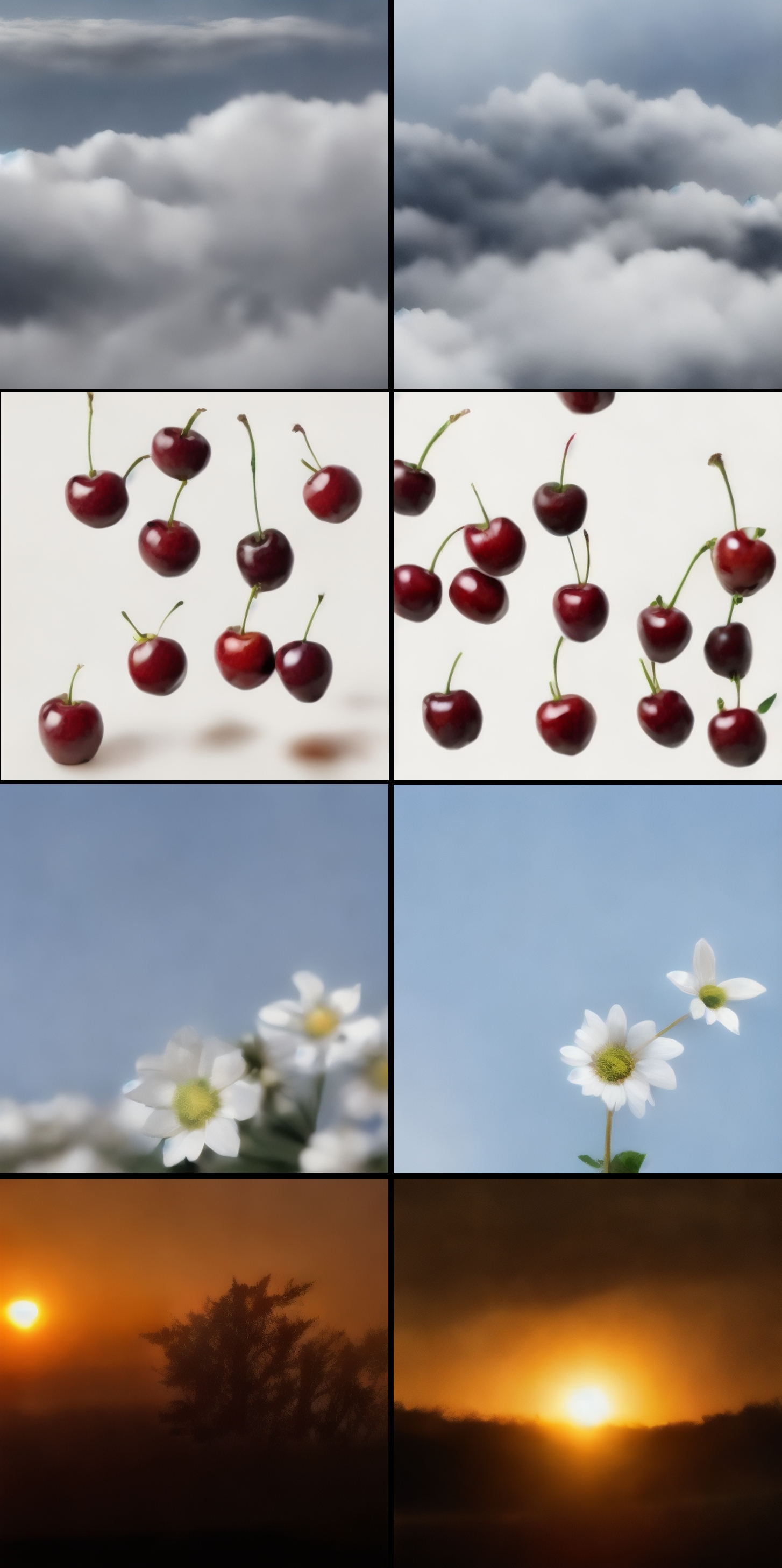}
    \caption{
Ablation demonstrating the effect of reintroducing the \texttt{E4--Mid--D4}
blocks into \texttt{NanoSD~1}. Column 1: NanoSD 1 and Column 2: NanoSD 1 + \texttt{E4--Mid--D4}. Although these additional blocks significantly increase the
parameter count (309\,M $\rightarrow$ 565\,M), the latency impact is minimal
(41\,ms $\rightarrow$ 46\,ms), since they operate at low spatial resolution.
However, as illustrated, the pre-finetuning image quality is nearly unchanged,
showing that the large memory overhead provides no practical benefit. This
justifies eliminating \texttt{E4--Mid--D4} from our search space for improved
edge efficiency.
}
    \label{fig:midblock}
\end{figure}

\begin{table*}[]
\centering
\caption{Quantitative comparison of different methods on DIV-2K Val \cite{div2kval}to  dataset. The best, second-best and third-best results are highlighted in red, blue, and green colors, respectively.}
\label{table:sr1}
\begin{tabular}{|l|l|l|l|l|l|l|l|l|l|}
\hline
Method                     & PSNR$\uparrow$  & SSIM$\uparrow$   & LPIPS$\downarrow$   & FID$\downarrow$  & NIQE$\downarrow$ & MUSIQ$\uparrow$ & Steps & MACs (G) & Para. (M) \\ \hline
StableSR \cite{stablesr}         & 23.26 & 0.572 & 0.311  & 24.44 & 4.75 & 65.92 & 200   & 79940    & 1410      \\ \hline
DiffBIR \cite{diffbir}                    & 23.64 & 0.564 & 0.352  & 30.72 & 4.70 & 65.81 & 50    & 24234    & 1717      \\ \hline
SeeSR \cite{seesr}                      & 23.68 & 0.604 & 0.319  & 25.90 & 4.81 & 68.67 & 50    & 65857    & 2524      \\ \hline
ResShift \cite{resshift}                   & \textcolor{red}{24.65} & \textcolor{blue}{0.618} & 0.334  & 36.11 & 6.82 & 61.09 & 15    & 5491     & 119       \\ \hline
SinSR \cite{sinsr}                     & \textcolor{blue}{24.41} & 0.601 & 0.324  & 35.57 & 6.02 & 62.82 & 1     & 2649     & 119       \\ \hline
OSEDiff \cite{osediff}                   & 23.72 & 0.610 & 0.294  & 26.32 & 4.71 & 67.97 & 1     & 2265     & 1775      \\ \hline
S3Diff \cite{s3diff}                     & 23.52 & 0.594 & \textcolor{blue}{0.258}  & \textcolor{red}{19.66} & 4.74 & 68.01  & 1     & 2621     & 1327      \\ \hline
TinySR \cite{tinysr}                       &   -    &    0.572    &   0.279         &  \textcolor{green}{22.94}    & \textcolor{blue}{4.15} & \textcolor{blue}{69.90}      &  1     &   427       &    341       \\ \hline
Edge-SD-SR \cite{edgesdsr}                  &   24.10    &  \textcolor{green}{0.617}      &  \textcolor{red}{0.249}            &   25.37    & - &  \textcolor{green}{69.58}     &    1   &    -      &    169     \\ \hline
PocketSR \cite{pocketsr}     & 23.85 & 0.601 & 0.280  & 25.25 & \textcolor{green}{4.415} & 66.38 & 1    & \textcolor{red}{225}    & 146      \\ \hline
AdcSR \cite{adcsr}                     & 23.74 & 0.601 & 0.285  & 25.52 & \textcolor{blue}{4.36} & 68.00 & 1     & 496      & 456       \\ \hline
 Nano-OSEDiff (Ours) &  \textcolor{green}{24.29}     &  \textcolor{red}{0.628}      &    0.296          &  27.46    & 4.92   &  66.41 &  1     &   \textcolor{green}{340}       &   448        \\ \hline
 Nano-S3Diff (Ours)  &  23.13     &  0.573      &       \textcolor{green}{0.278}         &  \textcolor{blue}{22.34}    &  \textcolor{red}{4.09}  &  \textcolor{red}{70.44} &    1   &   \textcolor{blue}{285}       &   318        \\ \hline
\end{tabular}
\end{table*}

\begin{table*}[]
\centering
\caption{Quantitative comparison of different methods on real-world datasets. The best, second-best and third-best results are highlighted in red, blue, and green colors, respectively.}
\label{table:sr2}
\begin{tabular}{|l|lllll|lllll|}
\hline
Datasets            & \multicolumn{5}{c|}{DRealSR \cite{drealsr}}                                                                                               & \multicolumn{5}{c|}{RealSR \cite{realsr}}                                                                                                \\ \hline
Method              & \multicolumn{1}{l|}{PSNR$\uparrow$}  & \multicolumn{1}{l|}{SSIM$\uparrow$}  & \multicolumn{1}{l|}{LPIPS$\downarrow$} & \multicolumn{1}{l|}{FID$\downarrow$}    & MUSIQ$\uparrow$ & \multicolumn{1}{l|}{PSNR$\uparrow$}  & \multicolumn{1}{l|}{SSIM$\uparrow$}  & \multicolumn{1}{l|}{LPIPS$\downarrow$} & \multicolumn{1}{l|}{FID$\downarrow$}    & MUSIQ$\uparrow$ \\ \hline
SinSR \cite{sinsr}               & \multicolumn{1}{l|}{\textcolor{blue}{28.36}} & \multicolumn{1}{l|}{0.751} & \multicolumn{1}{l|}{0.366} & \multicolumn{1}{l|}{170.5} & 55.33 & \multicolumn{1}{l|}{\textcolor{red}{26.28}} & \multicolumn{1}{l|}{\textcolor{green}{0.734}} & \multicolumn{1}{l|}{0.318} & \multicolumn{1}{l|}{135.9} & 60.80 \\ \hline
OSEDiff \cite{osediff}            & \multicolumn{1}{l|}{27.92} & \multicolumn{1}{l|}{\textcolor{blue}{0.783}} & \multicolumn{1}{l|}{0.296} & \multicolumn{1}{l|}{135.3} & 64.65 & \multicolumn{1}{l|}{25.15} & \multicolumn{1}{l|}{\textcolor{blue}{0.734}} & \multicolumn{1}{l|}{0.292} & \multicolumn{1}{l|}{123.4} & 69.09 \\ \hline
S3Diff \cite{s3diff}             & \multicolumn{1}{l|}{27.39} & \multicolumn{1}{l|}{0.746} & \multicolumn{1}{l|}{0.312} & \multicolumn{1}{l|}{\textcolor{red}{119.2}} & 64.16 & \multicolumn{1}{l|}{25.19} & \multicolumn{1}{l|}{0.731} & \multicolumn{1}{l|}{\textcolor{red}{0.270}} & \multicolumn{1}{l|}{\textcolor{red}{110.3}} & 67.92 \\ \hline
Edge-SD-SR \cite{edgesdsr}          & \multicolumn{1}{c|}{-}     & \multicolumn{1}{c|}{-}     & \multicolumn{1}{l|}{\textcolor{blue}{0.292}} & \multicolumn{1}{c|}{-}      & 55.66 & \multicolumn{1}{c|}{-}     & \multicolumn{1}{c|}{-}     & \multicolumn{1}{l|}{0.278} & \multicolumn{1}{c|}{-}      & 65.20 \\ \hline
AdcSR \cite{adcsr}              & \multicolumn{1}{l|}{\textcolor{green}{28.10}} & \multicolumn{1}{l|}{\textcolor{green}{0.772}} & \multicolumn{1}{l|}{0.304} & \multicolumn{1}{l|}{\textcolor{green}{134.0}} & \textcolor{blue}{66.26} & \multicolumn{1}{l|}{25.47} & \multicolumn{1}{l|}{0.730} & \multicolumn{1}{l|}{0.288} & \multicolumn{1}{l|}{118.4} & \textcolor{blue}{69.90} \\ \hline
TinySR \cite{tinysr}              & \multicolumn{1}{l|}{27.48} & \multicolumn{1}{l|}{0.745} & \multicolumn{1}{l|}{0.311} & \multicolumn{1}{l|}{146.7} & \textcolor{green}{65.36} & \multicolumn{1}{l|}{24.79} & \multicolumn{1}{l|}{0.717} & \multicolumn{1}{l|}{0.280} & \multicolumn{1}{l|}{\textcolor{green}{118}}    & \textcolor{green}{69.78} \\ \hline
PocketSR \cite{pocketsr}           & \multicolumn{1}{l|}{28.05} & \multicolumn{1}{l|}{0.767} & \multicolumn{1}{l|}{\textcolor{green}{0.296}} & \multicolumn{1}{c|}{-}      & 63.85 & \multicolumn{1}{l|}{\textcolor{green}{25.47}} & \multicolumn{1}{l|}{0.733} & \multicolumn{1}{l|}{\textcolor{blue}{0.271}} & \multicolumn{1}{c|}{-}      & 67.07 \\ \hline
Nano-OSEDiff  & \multicolumn{1}{l|}{\textcolor{red}{29.01}}      & \multicolumn{1}{l|}{\textcolor{red}{0.808}}      & \multicolumn{1}{l|}{\textcolor{red}{0.276}}      & \multicolumn{1}{l|}{134.6}       & 58.84      & \multicolumn{1}{l|}{\textcolor{blue}{26.19}}      & \multicolumn{1}{l|}{\textcolor{red}{0.746}}      & \multicolumn{1}{l|}{\textcolor{green}{0.272}}      & \multicolumn{1}{l|}{122.7}       &    65.29   \\ \hline
Nano-S3Diff   & \multicolumn{1}{l|}{26.96}      & \multicolumn{1}{l|}{0.735}      & \multicolumn{1}{l|}{0.321}      & \multicolumn{1}{l|}{\textcolor{blue}{127.1}}       &  \textcolor{red}{67.83}     & \multicolumn{1}{l|}{24.47}      & \multicolumn{1}{l|}{0.701}      & \multicolumn{1}{l|}{0.281}      & \multicolumn{1}{l|}{\textcolor{blue}{116.8}}       & \textcolor{red}{70.27}     \\ \hline
\end{tabular}
\end{table*}

\subsection{Ablation on Removing \texttt{E4--Mid--D4} Blocks.}
Prior work has shown that the highest-resolution stages of the U-Net dominate
per-layer computational cost, whereas the lowest-resolution stages contribute
disproportionately to the memory footprint with a limited effect on perceptual
quality. Motivated by this observation, our default search space removes the
encoder-4, mid, and decoder-4 blocks (\texttt{E4--Mid--D4}). To validate this
choice, we reintroduced these three blocks (configured as
\texttt{E4: R} $\rightarrow$ \texttt{Mid: RAR} $\rightarrow$ \texttt{D4: RR})
into one of our baseline models, \texttt{NanoSD~1} (see Table~1 of the main
paper). The original model has a latency of 41\,ms and 309\,M parameters. After
adding \texttt{E4--Mid--D4}, the parameter count increases substantially to
565\,M, while latency increases only marginally to 46\,ms due to the small
spatial resolutions processed by these blocks. 

We compare the pre-finetuning image quality of the two variants in
Figure~\ref{fig:midblock}. Despite the large increase in parameters and memory footprint,
the improvement in visual quality is negligible. This confirms that the
cost of including \texttt{E4--Mid--D4} is not justified for edge-efficient
models, and supports our decision to exclude these blocks from the search
space.

\subsection{Edge Efficient OSEDiff and S3Diff for Single Image Super-resolution}
 This paper aims to enable real-time super-resolution by effectively harnessing the rich generative priors inherent in pre-trained diffusion models. We validate the efficacy of the proposed foundation model for edge-efficient SR tasks by integrating NanoSD with prominent one-step latent diffusion frameworks, namely OSEDiff and S3Diff.
 
 OSEDiff employs LoRA layers integrated into the VAE encoder and diffusion UNet to perform super-resolution, thereby preserving the prior knowledge of the T2I model. We substitute the latent diffusion model in OSEDiff with NanoSD and fine-tune the asymmetric VAE encoder and UNet using LoRA, setting the rank to 4. Training follows the VSD framework consistent with OSEDiff. In contrast, S3Diff incorporates a pre-trained degradation estimation network to leverage rich content information from LR images. Its diffusion UNet is trained with a degradation-guided LoRA module and adversarial distillation for one-step SR. We integrate NanoSD into S3Diff, configuring rank parameters for the VAE encoder and diffusion UNet as 16 and 32, respectively, and adhere to S3Diff’s training protocols.

Table~\ref{table:sr1} and Table~\ref{table:sr2} present quantitative comparisons of super-resolution methods across three datasets. Both proposed models demonstrate robust performance and computational efficiency. Specifically, they exhibit the second-lowest MACs among evaluated approaches, enabling real-time processing and edge deployment. Nano-S3Diff achieves top NIQE and MUSIQ scores, along with second-best FID results, reflecting superior perceptual quality. Meanwhile, Nano-OSEDiff maintains competitive fidelity, yielding the superior PSNR and SSIM performance over Edge-SD-SR, PocketSR, and AdcSR while surpassing more computationally intensive methods. 

\begin{figure*}
    \centering
    \includegraphics[width=0.8\linewidth]{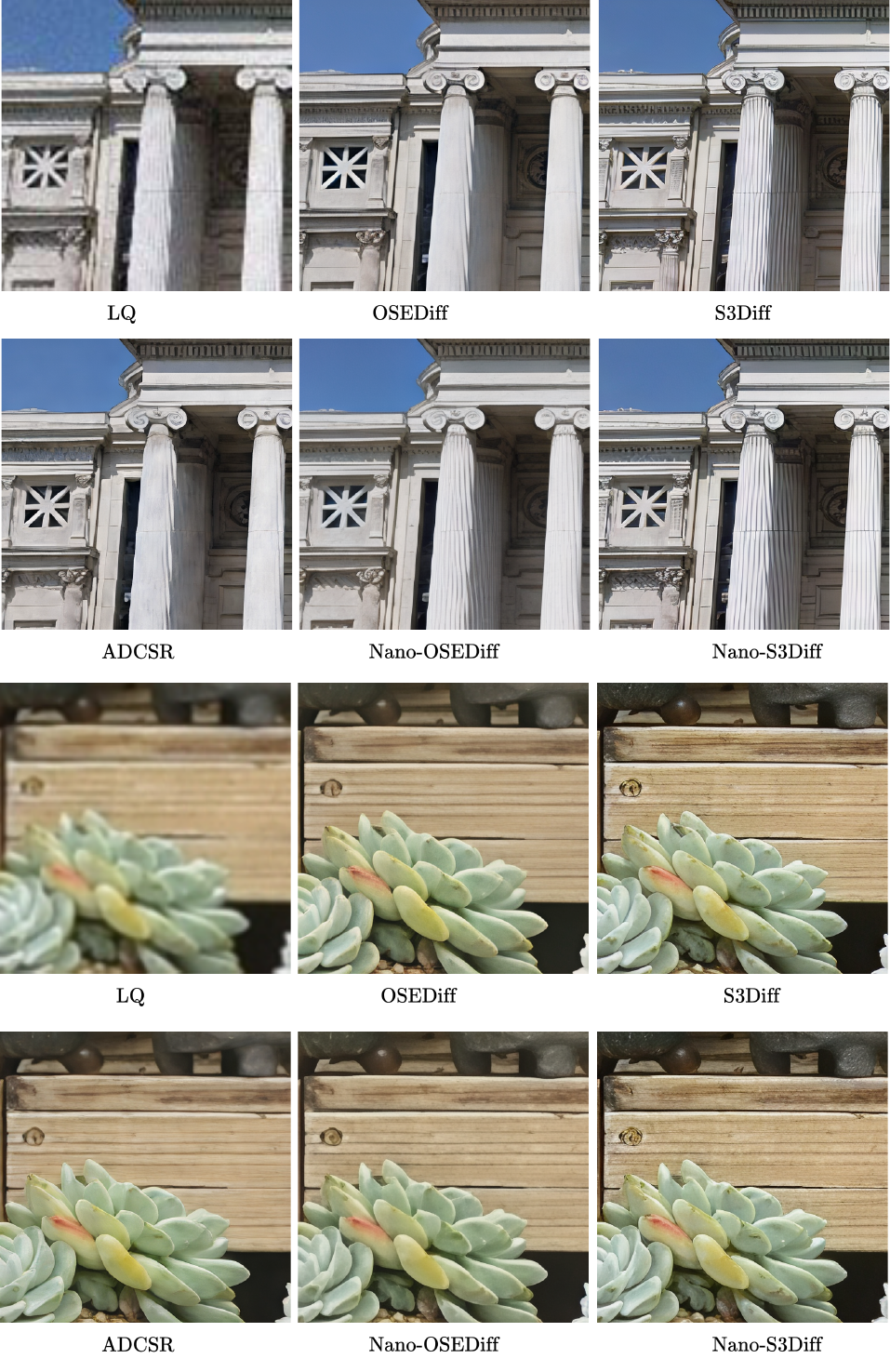}
    \caption{Qualitative comparisons of different SR methods. Please zoom in for a better view}
    \label{fig:sr1}
\end{figure*}

\begin{figure*}
    \centering
    \includegraphics[width=0.8\linewidth]{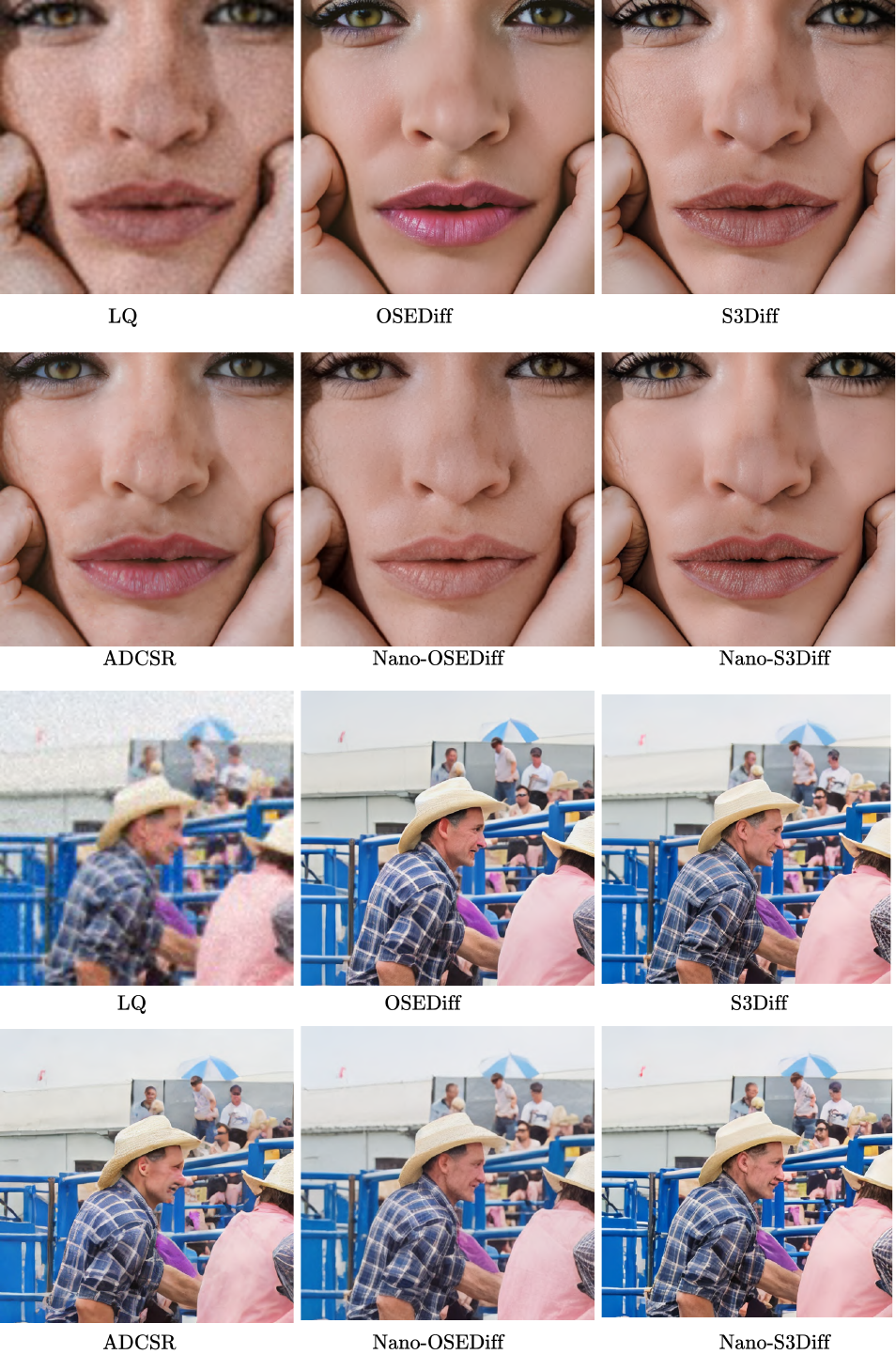}
    \caption{Qualitative comparisons of different SR methods. Please zoom in for a better view}
    \label{fig:sr2}
\end{figure*}

Fig.~\ref{fig:sr1} and Fig.~\ref{fig:sr2} present qualitative comparisons between different super-resolution methods. Our proposed approaches demonstrate consistent performance in preserving image fidelity and reconstructing fine details. In the first example, both Nano-OSEDiff and Nano-S3Diff accurately reconstruct the striped pattern, while AdCSR generates artifacts and unrealistic textures. The second example shows that our methods effectively restore person details, outperforming AdcSR in terms of both face restoration, structure recovery and detail preservation. These visual results confirm that our edge-efficient models achieve high-quality reconstructions while maintaining rich textural details across various image contexts.

\begin{table*}[]
\centering
\caption{Quantitative comparison of different methods on various real-world datasets. The best, second-best and third-best results are highlighted in red, blue, and green colors, respectively.}
\label{table:fr2}
\begin{tabular}{|l|lll|lll|lll|}
\hline
\multirow{2}{*}{Method} & \multicolumn{3}{l|}{Wider-Test \cite{codeformer}}                                & \multicolumn{3}{l|}{LFW-Test \cite{lfw}}                                  & \multicolumn{3}{l|}{WebPhoto-Test \cite{webphoto}}                             \\ \cline{2-10} 
                        & \multicolumn{1}{l|}{MUSIQ$\uparrow$} & \multicolumn{1}{l|}{NIQE$\downarrow$} & FID$\downarrow$   & \multicolumn{1}{l|}{MUSIQ$\uparrow$} & \multicolumn{1}{l|}{NIQE$\downarrow$} & FID$\downarrow$   & \multicolumn{1}{l|}{MUSIQ$\uparrow$} & \multicolumn{1}{l|}{NIQE$\downarrow$} & FID$\downarrow$   \\ \hline
PGDiff  \cite{pgdiff}                & \multicolumn{1}{l|}{68.13} & \multicolumn{1}{l|}{\textcolor{blue}{3.93}} & 35.86 & \multicolumn{1}{l|}{71.24} & \multicolumn{1}{l|}{4.01} & \textcolor{green}{41.20} & \multicolumn{1}{l|}{68.59} & \multicolumn{1}{l|}{\textcolor{blue}{3.99}} & \textcolor{green}{86.95} \\ \hline
DifFace \cite{difface}                & \multicolumn{1}{l|}{64.90} & \multicolumn{1}{l|}{4.23} & 37.09 & \multicolumn{1}{l|}{69.61} & \multicolumn{1}{l|}{\textcolor{green}{3.90}} & 46.12 & \multicolumn{1}{l|}{65.11} & \multicolumn{1}{l|}{4.24} & 79.55 \\ \hline
DiffBIR \cite{diffbir}                & \multicolumn{1}{l|}{\textcolor{red}{75.32}} & \multicolumn{1}{l|}{5.59} & 35.34 & \multicolumn{1}{l|}{\textcolor{red}{76.42}} & \multicolumn{1}{l|}{5.67} & \textcolor{red}{40.32} & \multicolumn{1}{l|}{72.27} & \multicolumn{1}{l|}{6.00} & \textcolor{red}{91.83} \\ \hline
OSEDiff \cite{osediff}                & \multicolumn{1}{l|}{70.55} & \multicolumn{1}{l|}{4.93} & 50.27 & \multicolumn{1}{l|}{73.40} & \multicolumn{1}{l|}{4.71} & 57.80 & \multicolumn{1}{l|}{72.59} & \multicolumn{1}{l|}{5.26} & 117.5 \\ \hline
OSDFace \cite{osdface}                & \multicolumn{1}{l|}{\textcolor{green}{74.60}} & \multicolumn{1}{l|}{\textcolor{red}{3.77}} & \textcolor{red}{34.64} & \multicolumn{1}{l|}{\textcolor{green}{75.35}} & \multicolumn{1}{l|}{\textcolor{red}{3.87}} & 51.04 & \multicolumn{1}{l|}{\textcolor{blue}{73.93}} & \multicolumn{1}{l|}{\textcolor{red}{3.98}} & 84.59 \\ \hline
Nano-DiffBIR (Ours)     & \multicolumn{1}{l|}{\textcolor{blue}{74.88}}      & \multicolumn{1}{l|}{5.73}     &   \textcolor{blue}{35.13}    & \multicolumn{1}{l|}{\textcolor{blue}{76.02}}      & \multicolumn{1}{l|}{5.81}     &  \textcolor{blue}{40.45}     & \multicolumn{1}{l|}{\textcolor{green}{72.99}}      & \multicolumn{1}{l|}{5.89}     &   \textcolor{blue}{90.48}   \\ \hline
Nano-OSDFace (Ours)     & \multicolumn{1}{l|}{74.32}      & \multicolumn{1}{l|}{\textcolor{green}{\textcolor{green}{3.96}}}     &  \textcolor{green}{35.19}     & \multicolumn{1}{l|}{75.28}      & \multicolumn{1}{l|}{\textcolor{blue}{3.81}}     &    53.21   & \multicolumn{1}{l|}{\textcolor{red}{74.96}}      & \multicolumn{1}{l|}{\textcolor{green}{4.01}}     &  85.03     \\ \hline
\end{tabular}
\end{table*}

\subsection{Edge Efficient OSDFace for Face Restoration}
This section evaluates the applicability of our proposed model for face restoration tasks. To demonstrate its effectiveness, we integrate NanoSD into the OSDFace framework, which represents the current state-of-the-art one-step diffusion approach for face restoration. 

\begin{figure*}
        \centering
        \includegraphics[width=0.85\linewidth]{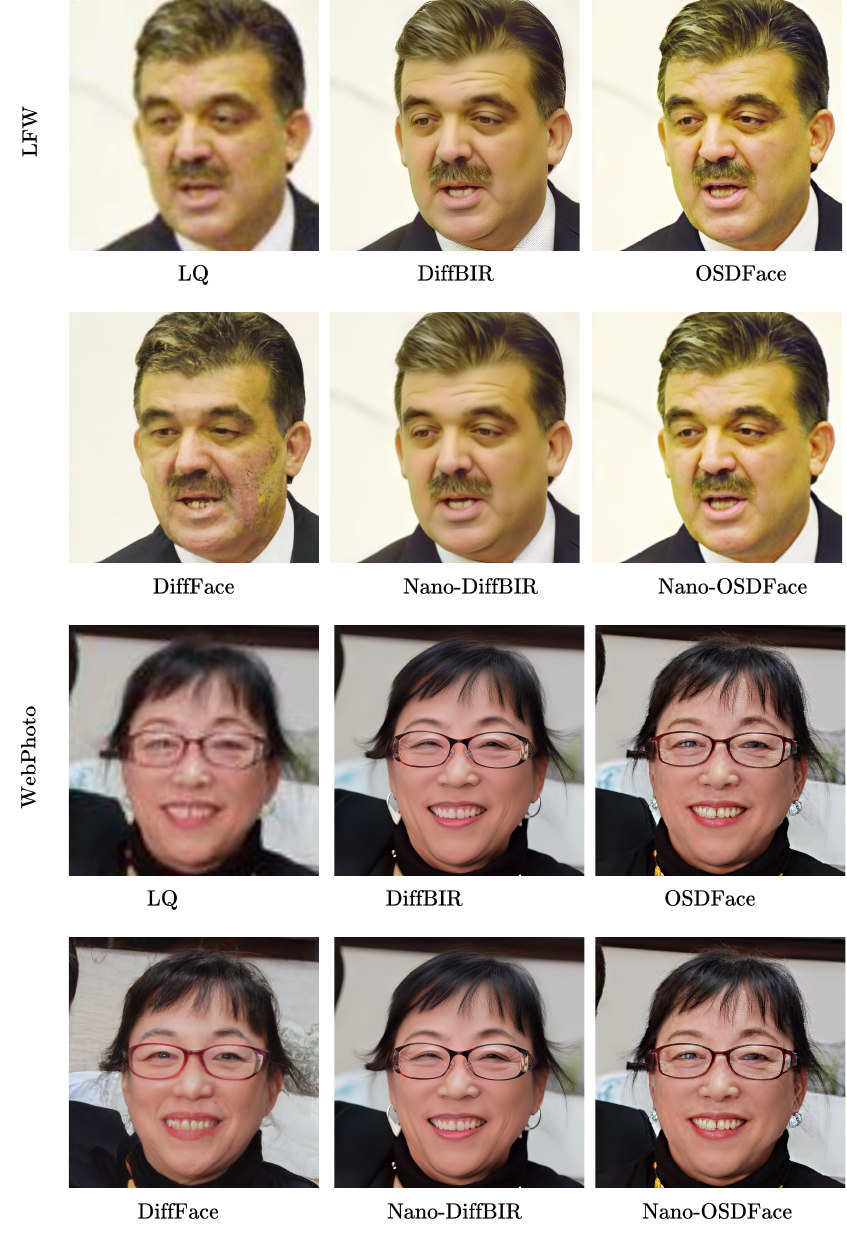}
        \caption{Qualitative comparisons of different FR methods.}
        \label{fig:fr1}
\end{figure*}

OSDFace acquires comprehensive facial prior features to guide the diffusion process. These priors are generated by training a Visual Representation Embedder (VRE) on low-quality (LQ) facial images. This prior knowledge is integrated into the diffusion UNet using cross-attention layers, with model fine-tuning accomplished via LoRA. To optimize inference speed, NanoSD is incorporated into the OSDFace framework. This implementation retains VRE for prompt embedding construction and employs LoRA-based UNet fine-tuning at a rank of 32. Table~\ref{table:fr2} and Fig.~\ref{fig:fr1} present  comparisons of various methods on three real-world datasets, where our method consistently delivers competitive results. These findings demonstrate that Nano-OSDFace achieves substantial improvements in efficiency while maintaining performance parity with leading face restoration approaches.

\subsection{Accelerating Diff-Plugin for Image Restoration with Generative Diffusion Prior}
\begin{figure*}
    \centering
    \includegraphics[width=1\linewidth]{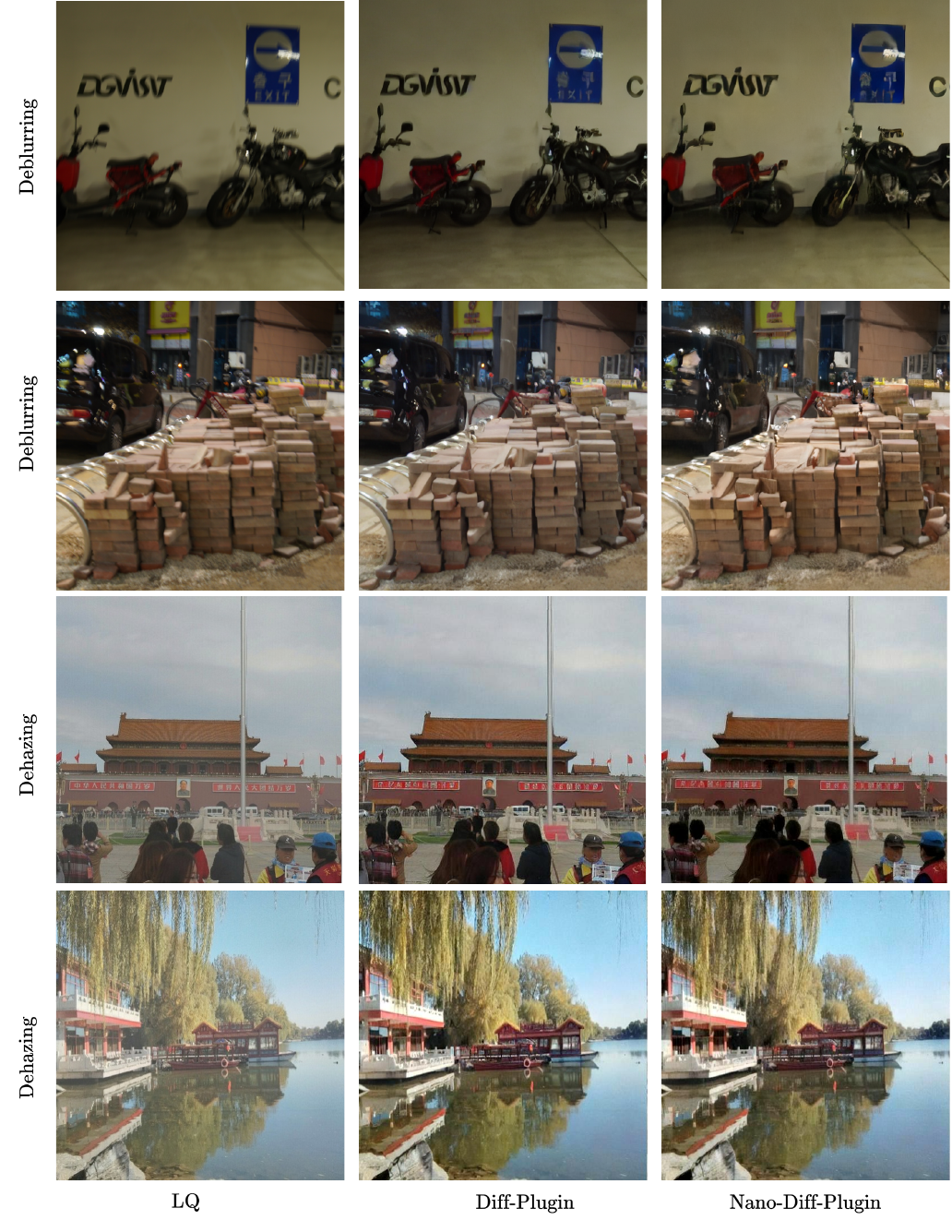}
    \caption{Qualitative comparisons of different methods for various
restoration tasks}
    \label{fig:diffplugin1}
\end{figure*}

\begin{figure*}
    \centering
    \includegraphics[width=1\linewidth]{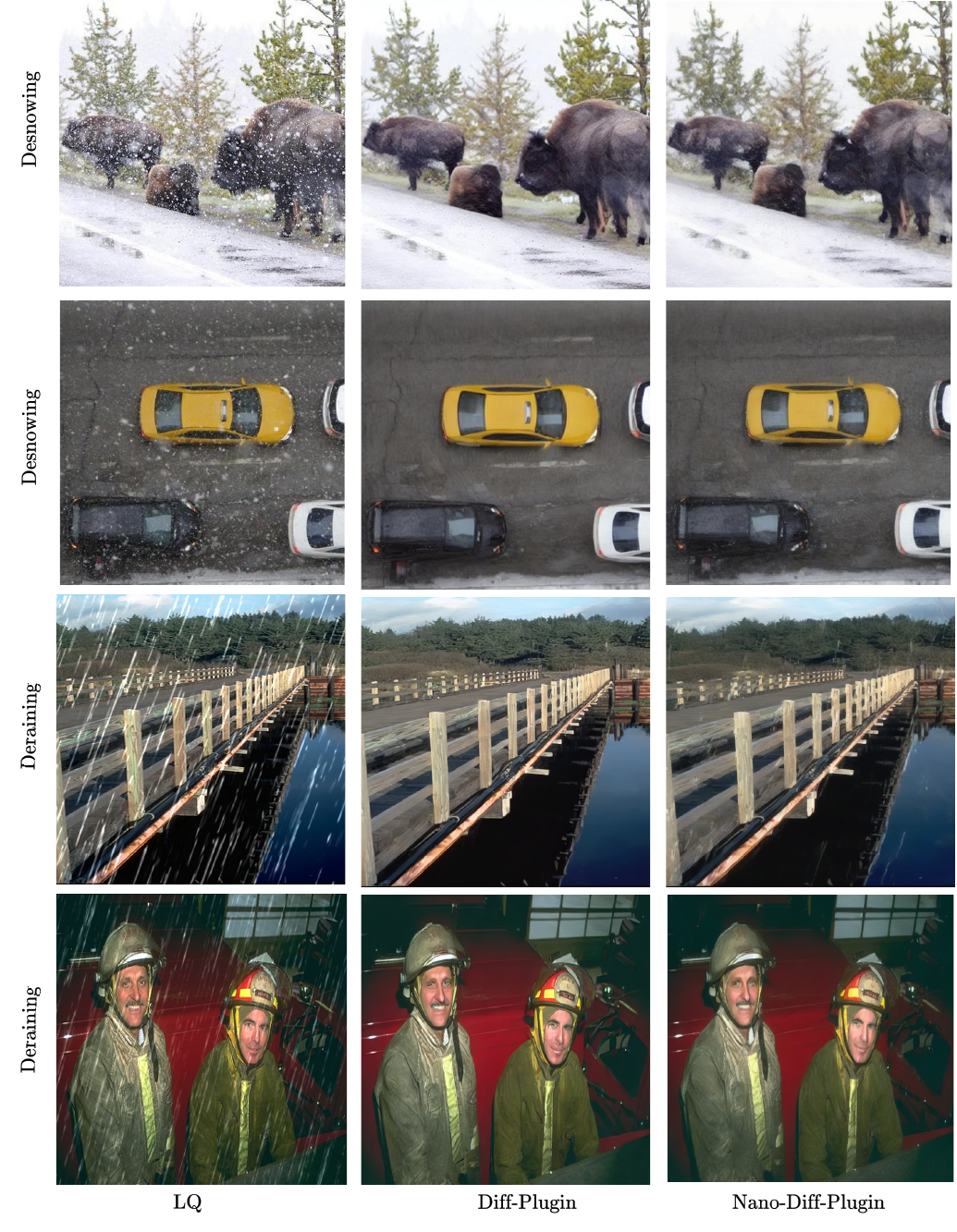}
    \caption{Qualitative comparisons of different methods for various
restoration tasks}
    \label{fig:diffplugin2}
\end{figure*}

The Diff-Plugin framework employs task-specific priors that combine task guidance and input image spatial information. These priors enable pre-trained diffusion models to address low-level vision tasks while preserving content fidelity. Central to this approach is a lightweight Task-Plugin module, which comprises two components: the Task-Prompt Branch (TPB) for task-specific direction and the Spatial Complement Branch (SCB) for enhancing output fidelity through visual guidance. In our implementation, we substitute the original latent diffusion backbone with NanoSD. Task-specific priors extracted by the Task-Plugin are injected into NanoSD's ResNet and Cross-Attention blocks. Similar to the standard Diff-Plugin framework, the module is optimized using a denoising loss to effectively integrate task-specific guidance into the diffusion process. Fig.~\ref{fig:diffplugin1} and Fig.~\ref{fig:diffplugin2} illustrate the enhanced performance of Nano-Diff-Plugin across four challenging low-level vision tasks. Visual comparisons reveal that our lightweight approach delivers consistent results across all evaluated tasks.

%% file: main.bib
@inproceedings{derain,
  title={Multi-stage progressive image restoration},
  author={Zamir, Syed Waqas and Arora, Aditya and Khan, Salman and Hayat, Munawar and Khan, Fahad Shahbaz and Yang, Ming-Hsuan and Shao, Ling},
  booktitle={Proceedings of the IEEE/CVF conference on computer vision and pattern recognition},
  pages={14821--14831},
  year={2021}
}

@article{rtts,
  title={Benchmarking single-image dehazing and beyond},
  author={Li, Boyi and Ren, Wenqi and Fu, Dengpan and Tao, Dacheng and Feng, Dan and Zeng, Wenjun and Wang, Zhangyang},
  journal={IEEE transactions on image processing},
  volume={28},
  number={1},
  pages={492--505},
  year={2018},
  publisher={IEEE}
}

@article{laion,
  title={Laion-5b: An open large-scale dataset for training next generation image-text models},
  author={Schuhmann, Christoph and Beaumont, Romain and Vencu, Richard and Gordon, Cade and Wightman, Ross and Cherti, Mehdi and Coombes, Theo and Katta, Aarush and Mullis, Clayton and Wortsman, Mitchell and others},
  journal={Advances in neural information processing systems},
  volume={35},
  pages={25278--25294},
  year={2022}
}

@article{tinysd,
  title={On architectural compression of text-to-image diffusion models},
  author={Kim, Bo-Kyeong and Song, Hyoung-Kyu and Castells, Thibault and Choi, Shinkook},
  year={2023}
}

@article{koala,
  title={Koala: Empirical lessons toward memory-efficient and fast diffusion models for text-to-image synthesis},
  author={Lee, Youngwan and Park, Kwanyong and Cho, Yoorhim and Lee, Yong-Ju and Hwang, Sung Ju},
  journal={Advances in Neural Information Processing Systems},
  volume={37},
  pages={51597--51633},
  year={2024}
}

@article{stablesr,
  title={Exploiting diffusion prior for real-world image super-resolution},
  author={Wang, Jianyi and Yue, Zongsheng and Zhou, Shangchen and Chan, Kelvin CK and Loy, Chen Change},
  journal={International Journal of Computer Vision},
  volume={132},
  number={12},
  pages={5929--5949},
  year={2024},
  publisher={Springer}
}

@inproceedings{diffpir,
  title={Denoising diffusion models for plug-and-play image restoration},
  author={Zhu, Yuanzhi and Zhang, Kai and Liang, Jingyun and Cao, Jiezhang and Wen, Bihan and Timofte, Radu and Van Gool, Luc},
  booktitle={Proceedings of the IEEE/CVF conference on computer vision and pattern recognition},
  pages={1219--1229},
  year={2023}
}

@inproceedings{sd,
  title={High-resolution image synthesis with latent diffusion models},
  author={Rombach, Robin and Blattmann, Andreas and Lorenz, Dominik and Esser, Patrick and Ommer, Bj{\"o}rn},
  booktitle={Proceedings of the IEEE/CVF conference on computer vision and pattern recognition},
  pages={10684--10695},
  year={2022}
}

@inproceedings{sd3,
  title={Scaling rectified flow transformers for high-resolution image synthesis},
  author={Esser, Patrick and Kulal, Sumith and Blattmann, Andreas and Entezari, Rahim and M{\"u}ller, Jonas and Saini, Harry and Levi, Yam and Lorenz, Dominik and Sauer, Axel and Boesel, Frederic and others},
  booktitle={Forty-first international conference on machine learning},
  year={2024}
}

@inproceedings{pasd,
  title={Pixel-aware stable diffusion for realistic image super-resolution and personalized stylization},
  author={Yang, Tao and Wu, Rongyuan and Ren, Peiran and Xie, Xuansong and Zhang, Lei},
  booktitle={European conference on computer vision},
  pages={74--91},
  year={2024},
  organization={Springer}
}

@article{snapfusion,
  title={Snapfusion: Text-to-image diffusion model on mobile devices within two seconds},
  author={Li, Yanyu and Wang, Huan and Jin, Qing and Hu, Ju and Chemerys, Pavlo and Fu, Yun and Wang, Yanzhi and Tulyakov, Sergey and Ren, Jian},
  journal={Advances in Neural Information Processing Systems},
  volume={36},
  pages={20662--20678},
  year={2023}
}

@inproceedings{mobilediffusion,
  title={Mobilediffusion: Instant text-to-image generation on mobile devices},
  author={Zhao, Yang and Xu, Yanwu and Xiao, Zhisheng and Jia, Haolin and Hou, Tingbo},
  booktitle={European Conference on Computer Vision},
  pages={225--242},
  year={2024},
  organization={Springer}
}

@inproceedings{snapgen,
  title={Snapgen: Taming high-resolution text-to-image models for mobile devices with efficient architectures and training},
  author={Chen, Jierun and Hu, Dongting and Huang, Xijie and Coskun, Huseyin and Sahni, Arpit and Gupta, Aarush and Goyal, Anujraaj and Lahiri, Dishani and Singh, Rajesh and Idelbayev, Yerlan and others},
  booktitle={Proceedings of the Computer Vision and Pattern Recognition Conference},
  pages={7997--8008},
  year={2025}
}

@inproceedings{diffbir,
  title={Diffbir: Toward blind image restoration with generative diffusion prior},
  author={Lin, Xinqi and He, Jingwen and Chen, Ziyan and Lyu, Zhaoyang and Dai, Bo and Yu, Fanghua and Qiao, Yu and Ouyang, Wanli and Dong, Chao},
  booktitle={European conference on computer vision},
  pages={430--448},
  year={2024},
  organization={Springer}
}

@inproceedings{seesr,
  title={Seesr: Towards semantics-aware real-world image super-resolution},
  author={Wu, Rongyuan and Yang, Tao and Sun, Lingchen and Zhang, Zhengqiang and Li, Shuai and Zhang, Lei},
  booktitle={Proceedings of the IEEE/CVF conference on computer vision and pattern recognition},
  pages={25456--25467},
  year={2024}
}

@article{resshift,
  title={Resshift: Efficient diffusion model for image super-resolution by residual shifting},
  author={Yue, Zongsheng and Wang, Jianyi and Loy, Chen Change},
  journal={Advances in Neural Information Processing Systems},
  volume={36},
  pages={13294--13307},
  year={2023}
}

@inproceedings{sinsr,
  title={Sinsr: diffusion-based image super-resolution in a single step},
  author={Wang, Yufei and Yang, Wenhan and Chen, Xinyuan and Wang, Yaohui and Guo, Lanqing and Chau, Lap-Pui and Liu, Ziwei and Qiao, Yu and Kot, Alex C and Wen, Bihan},
  booktitle={Proceedings of the IEEE/CVF conference on computer vision and pattern recognition},
  pages={25796--25805},
  year={2024}
}

@article{osediff,
  title={One-step effective diffusion network for real-world image super-resolution},
  author={Wu, Rongyuan and Sun, Lingchen and Ma, Zhiyuan and Zhang, Lei},
  journal={Advances in Neural Information Processing Systems},
  volume={37},
  pages={92529--92553},
  year={2024}
}

@article{s3diff,
  title={Degradation-guided one-step image super-resolution with diffusion priors},
  author={Zhang, Aiping and Yue, Zongsheng and Pei, Renjing and Ren, Wenqi and Cao, Xiaochun},
  journal={arXiv preprint arXiv:2409.17058},
  year={2024}
}

@article{tinysr,
  title={TinySR: Pruning Diffusion for Real-World Image Super-Resolution},
  author={Dong, Linwei and Fan, Qingnan and Yu, Yuhang and Zhang, Qi and Chen, Jinwei and Luo, Yawei and Zou, Changqing},
  journal={arXiv preprint arXiv:2508.17434},
  year={2025}
}

@article{pocketsr,
  title={PocketSR: The Super-Resolution Expert in Your Pocket Mobiles},
  author={Sun, Haoze and Jiang, Linfeng and Li, Fan and Pei, Renjing and Wang, Zhixin and Guo, Yong and Xu, Jiaqi and Chen, Haoyu and Han, Jin and Song, Fenglong and others},
  journal={arXiv preprint arXiv:2510.03012},
  year={2025}
}

@inproceedings{edgesdsr,
  title={Edge-SD-SR: Low Latency and Parameter Efficient On-device Super-Resolution with Stable Diffusion via Bidirectional Conditioning},
  author={Hadji, Isma and Noroozi, Mehdi and Escorcia, Victor and Zaganidis, Anestis and Martinez, Brais and Tzimiropoulos, Georgios},
  booktitle={Proceedings of the Computer Vision and Pattern Recognition Conference},
  pages={12789--12798},
  year={2025}
}

@inproceedings{adcsr,
  title={Adversarial diffusion compression for real-world image super-resolution},
  author={Chen, Bin and Li, Gehui and Wu, Rongyuan and Zhang, Xindong and Chen, Jie and Zhang, Jian and Zhang, Lei},
  booktitle={Proceedings of the Computer Vision and Pattern Recognition Conference},
  pages={28208--28220},
  year={2025}
}

@article{pgdiff,
  title={PGDiff: Guiding diffusion models for versatile face restoration via partial guidance},
  author={Yang, Peiqing and Zhou, Shangchen and Tao, Qingyi and Loy, Chen Change},
  journal={Advances in Neural Information Processing Systems},
  volume={36},
  pages={32194--32214},
  year={2023}
}

@article{difface,
  title={Difface: Blind face restoration with diffused error contraction},
  author={Yue, Zongsheng and Loy, Chen Change},
  journal={IEEE Transactions on Pattern Analysis and Machine Intelligence},
  year={2024},
  publisher={IEEE}
}

@inproceedings{osdface,
  title={Osdface: One-step diffusion model for face restoration},
  author={Wang, Jingkai and Gong, Jue and Zhang, Lin and Chen, Zheng and Liu, Xing and Gu, Hong and Liu, Yutong and Zhang, Yulun and Yang, Xiaokang},
  booktitle={Proceedings of the Computer Vision and Pattern Recognition Conference},
  pages={12626--12636},
  year={2025}
}

@article{codeformer,
  title={Towards robust blind face restoration with codebook lookup transformer},
  author={Zhou, Shangchen and Chan, Kelvin and Li, Chongyi and Loy, Chen Change},
  journal={Advances in Neural Information Processing Systems},
  volume={35},
  pages={30599--30611},
  year={2022}
}

@inproceedings{instructp2p,
  title={Instructpix2pix: Learning to follow image editing instructions},
  author={Brooks, Tim and Holynski, Aleksander and Efros, Alexei A},
  booktitle={Proceedings of the IEEE/CVF conference on computer vision and pattern recognition},
  pages={18392--18402},
  year={2023}
}

@inproceedings{nulltext,
  title={Null-text inversion for editing real images using guided diffusion models},
  author={Mokady, Ron and Hertz, Amir and Aberman, Kfir and Pritch, Yael and Cohen-Or, Daniel},
  booktitle={Proceedings of the IEEE/CVF conference on computer vision and pattern recognition},
  pages={6038--6047},
  year={2023}
}

@inproceedings{controlnet,
  title={Adding conditional control to text-to-image diffusion models},
  author={Zhang, Lvmin and Rao, Anyi and Agrawala, Maneesh},
  booktitle={Proceedings of the IEEE/CVF international conference on computer vision},
  pages={3836--3847},
  year={2023}
}

@inproceedings{diffplugin,
  title={Diff-plugin: Revitalizing details for diffusion-based low-level tasks},
  author={Liu, Yuhao and Ke, Zhanghan and Liu, Fang and Zhao, Nanxuan and Lau, Rynson WH},
  booktitle={Proceedings of the IEEE/CVF Conference on Computer Vision and Pattern Recognition},
  pages={4197--4208},
  year={2024}
}

@article{realistic,
  title={Desnownet: Context-aware deep network for snow removal},
  author={Liu, Yun-Fu and Jaw, Da-Wei and Huang, Shih-Chia and Hwang, Jenq-Neng},
  journal={IEEE Transactions on Image Processing},
  volume={27},
  number={6},
  pages={3064--3073},
  year={2018},
  publisher={IEEE}
}

@article{reside,
  title={Benchmarking single-image dehazing and beyond},
  author={Li, Boyi and Ren, Wenqi and Fu, Dengpan and Tao, Dacheng and Feng, Dan and Zeng, Wenjun and Wang, Zhangyang},
  journal={IEEE transactions on image processing},
  volume={28},
  number={1},
  pages={492--505},
  year={2018},
  publisher={IEEE}
}

@inproceedings{realblurj,
  title={Real-world blur dataset for learning and benchmarking deblurring algorithms},
  author={Rim, Jaesung and Lee, Haeyun and Won, Jucheol and Cho, Sunghyun},
  booktitle={European conference on computer vision},
  pages={184--201},
  year={2020},
  organization={Springer}
}

@inproceedings{realtest,
  title={Spatial attentive single-image deraining with a high quality real rain dataset},
  author={Wang, Tianyu and Yang, Xin and Xu, Ke and Chen, Shaozhe and Zhang, Qiang and Lau, Rynson WH},
  booktitle={Proceedings of the IEEE/CVF conference on computer vision and pattern recognition},
  pages={12270--12279},
  year={2019}
}

@inproceedings{lfw,
  title={Labeled faces in the wild: A database forstudying face recognition in unconstrained environments},
  author={Huang, Gary B and Mattar, Marwan and Berg, Tamara and Learned-Miller, Eric},
  booktitle={Workshop on faces in'Real-Life'Images: detection, alignment, and recognition},
  year={2008}
}

@inproceedings{webphoto,
  title={Towards real-world blind face restoration with generative facial prior},
  author={Wang, Xintao and Li, Yu and Zhang, Honglun and Shan, Ying},
  booktitle={Proceedings of the IEEE/CVF conference on computer vision and pattern recognition},
  pages={9168--9178},
  year={2021}
}

@article{celebtest,
  title={Progressive growing of gans for improved quality, stability, and variation},
  author={Karras, Tero and Aila, Timo and Laine, Samuli and Lehtinen, Jaakko},
  journal={arXiv preprint arXiv:1710.10196},
  year={2017}
}

@inproceedings{div2kval,
  title={Ntire 2017 challenge on single image super-resolution: Dataset and study},
  author={Agustsson, Eirikur and Timofte, Radu},
  booktitle={Proceedings of the IEEE conference on computer vision and pattern recognition workshops},
  pages={126--135},
  year={2017}
}

@inproceedings{realsr,
  title={Toward real-world single image super-resolution: A new benchmark and a new model},
  author={Cai, Jianrui and Zeng, Hui and Yong, Hongwei and Cao, Zisheng and Zhang, Lei},
  booktitle={Proceedings of the IEEE/CVF international conference on computer vision},
  pages={3086--3095},
  year={2019}
}

@inproceedings{drealsr,
  title={Component divide-and-conquer for real-world image super-resolution},
  author={Wei, Pengxu and Xie, Ziwei and Lu, Hannan and Zhan, Zongyuan and Ye, Qixiang and Zuo, Wangmeng and Lin, Liang},
  booktitle={European conference on computer vision},
  pages={101--117},
  year={2020},
  organization={Springer}
}

@inproceedings{lsdir,
  title={Lsdir: A large scale dataset for image restoration},
  author={Li, Yawei and Zhang, Kai and Liang, Jingyun and Cao, Jiezhang and Liu, Ce and Gong, Rui and Zhang, Yulun and Tang, Hao and Liu, Yun and Demandolx, Denis and others},
  booktitle={Proceedings of the IEEE/CVF Conference on Computer Vision and Pattern Recognition},
  pages={1775--1787},
  year={2023}
}

@inproceedings{lpips,
  title={The unreasonable effectiveness of deep features as a perceptual metric},
  author={Zhang, Richard and Isola, Phillip and Efros, Alexei A and Shechtman, Eli and Wang, Oliver},
  booktitle={Proceedings of the IEEE conference on computer vision and pattern recognition},
  pages={586--595},
  year={2018}
}

@article{fid,
  title={Gans trained by a two time-scale update rule converge to a local nash equilibrium},
  author={Heusel, Martin and Ramsauer, Hubert and Unterthiner, Thomas and Nessler, Bernhard and Hochreiter, Sepp},
  journal={Advances in neural information processing systems},
  volume={30},
  year={2017}
}

@article{niqe,
  title={A feature-enriched completely blind image quality evaluator},
  author={Zhang, Lin and Zhang, Lei and Bovik, Alan C},
  journal={IEEE Transactions on Image Processing},
  volume={24},
  number={8},
  pages={2579--2591},
  year={2015},
  publisher={IEEE}
}

@inproceedings{musiq,
  title={Musiq: Multi-scale image quality transformer},
  author={Ke, Junjie and Wang, Qifei and Wang, Yilin and Milanfar, Peyman and Yang, Feng},
  booktitle={Proceedings of the IEEE/CVF international conference on computer vision},
  pages={5148--5157},
  year={2021}
}

@article{ssim,
  title={Image quality assessment: from error visibility to structural similarity},
  author={Wang, Zhou and Bovik, Alan C and Sheikh, Hamid R and Simoncelli, Eero P},
  journal={IEEE transactions on image processing},
  volume={13},
  number={4},
  pages={600--612},
  year={2004},
  publisher={IEEE}
}

@article{dist,
  title={Image quality assessment: Unifying structure and texture similarity},
  author={Ding, Keyan and Ma, Kede and Wang, Shiqi and Simoncelli, Eero P},
  journal={IEEE transactions on pattern analysis and machine intelligence},
  volume={44},
  number={5},
  pages={2567--2581},
  year={2020},
  publisher={IEEE}
}

@inproceedings{ffhq,
  title={A style-based generator architecture for generative adversarial networks},
  author={Karras, Tero and Laine, Samuli and Aila, Timo},
  booktitle={Proceedings of the IEEE/CVF conference on computer vision and pattern recognition},
  pages={4401--4410},
  year={2019}
}

@inproceedings{gopro,
  title={Deep multi-scale convolutional neural network for dynamic scene deblurring},
  author={Nah, Seungjun and Hyun Kim, Tae and Mu Lee, Kyoung},
  booktitle={Proceedings of the IEEE conference on computer vision and pattern recognition},
  pages={3883--3891},
  year={2017}
}

@article{snow100k,
  title={Desnownet: Context-aware deep network for snow removal},
  author={Liu, Yun-Fu and Jaw, Da-Wei and Huang, Shih-Chia and Hwang, Jenq-Neng},
  journal={IEEE Transactions on Image Processing},
  volume={27},
  number={6},
  pages={3064--3073},
  year={2018},
  publisher={IEEE}
}

@inproceedings{marigold,
  title={Repurposing diffusion-based image generators for monocular depth estimation},
  author={Ke, Bingxin and Obukhov, Anton and Huang, Shengyu and Metzger, Nando and Daudt, Rodrigo Caye and Schindler, Konrad},
  booktitle={Proceedings of the IEEE/CVF conference on computer vision and pattern recognition},
  pages={9492--9502},
  year={2024}
}

@inproceedings{hypersim,
  title={Hypersim: A photorealistic synthetic dataset for holistic indoor scene understanding},
  author={Roberts, Mike and Ramapuram, Jason and Ranjan, Anurag and Kumar, Atulit and Bautista, Miguel Angel and Paczan, Nathan and Webb, Russ and Susskind, Joshua M},
  booktitle={Proceedings of the IEEE/CVF international conference on computer vision},
  pages={10912--10922},
  year={2021}
}

@article{virtualkitti,
  title={Virtual kitti 2},
  author={Cabon, Yohann and Murray, Naila and Humenberger, Martin},
  journal={arXiv preprint arXiv:2001.10773},
  year={2020}
}

@inproceedings{nyv2,
  title={Indoor segmentation and support inference from rgbd images},
  author={Silberman, Nathan and Hoiem, Derek and Kohli, Pushmeet and Fergus, Rob},
  booktitle={European conference on computer vision},
  pages={746--760},
  year={2012},
  organization={Springer}
}

@inproceedings{kitti,
  title={Are we ready for autonomous driving? the kitti vision benchmark suite},
  author={Geiger, Andreas and Lenz, Philip and Urtasun, Raquel},
  booktitle={2012 IEEE conference on computer vision and pattern recognition},
  pages={3354--3361},
  year={2012},
  organization={IEEE}
}

@article{midasabsrel,
  title={Towards robust monocular depth estimation: Mixing datasets for zero-shot cross-dataset transfer},
  author={Ranftl, Ren{\'e} and Lasinger, Katrin and Hafner, David and Schindler, Konrad and Koltun, Vladlen},
  journal={IEEE transactions on pattern analysis and machine intelligence},
  volume={44},
  number={3},
  pages={1623--1637},
  year={2020},
  publisher={IEEE}
}

@inproceedings{dptdelta,
  title={Vision transformers for dense prediction},
  author={Ranftl, Ren{\'e} and Bochkovskiy, Alexey and Koltun, Vladlen},
  booktitle={Proceedings of the IEEE/CVF international conference on computer vision},
  pages={12179--12188},
  year={2021}
}

@inproceedings{lerecover,
  title={Learning to recover 3d scene shape from a single image},
  author={Yin, Wei and Zhang, Jianming and Wang, Oliver and Niklaus, Simon and Mai, Long and Chen, Simon and Shen, Chunhua},
  booktitle={Proceedings of the IEEE/CVF conference on computer vision and pattern recognition},
  pages={204--213},
  year={2021}
}

@inproceedings{omnidata,
  title={Omnidata: A scalable pipeline for making multi-task mid-level vision datasets from 3d scans},
  author={Eftekhar, Ainaz and Sax, Alexander and Malik, Jitendra and Zamir, Amir},
  booktitle={Proceedings of the IEEE/CVF International Conference on Computer Vision},
  pages={10786--10796},
  year={2021}
}

@article{hdn,
  title={Hierarchical normalization for robust monocular depth estimation},
  author={Zhang, Chi and Yin, Wei and Wang, Billzb and Yu, Gang and Fu, Bin and Shen, Chunhua},
  journal={Advances in Neural Information Processing Systems},
  volume={35},
  pages={14128--14139},
  year={2022}
}

@inproceedings{snoek2012practical,
  title={Practical Bayesian Optimization of Machine Learning Algorithms},
  author={Snoek, Jasper and Larochelle, Hugo and Adams, Ryan P.},
  booktitle={NeurIPS},
  year={2012}
}
